\documentclass[oneref]{urithesis}
\usepackage{amsmath,amsthm, amsfonts, amssymb, amsxtra,amsopn}
\usepackage{graphicx}
\usepackage{multirow}
\usepackage{longtable}
\usepackage{booktabs}
\usepackage{cmap}
\PassOptionsToPackage{hyphens}{url}
\usepackage{xcolor, colortbl} % Used for coloring the cells of tables
\usepackage{amsmath}
\usepackage[T1]{fontenc}
\usepackage{textcomp} % Required for upquote.
\usepackage{listings} % Used for printing source code in papers
\usepackage{microtype}
\usepackage{pgfplots} % Used to create graphs
\usepackage{makecell} % Used to create thick links in tables
\usepackage[shortcuts]{extdash} % Allow the use of unbreakable "extdash" spacers.
\usepackage{subcaption}

\usepackage{enumitem}
\setlist[itemize]{noitemsep, topsep=0pt}

\usepackage{algorithm} % Used for writing algorithms in a paper
\usepackage[noend]{algpseudocode} % Allows psuedocode keywords (e.g., "if", "while", "for", etc.) in algorithms.

\usepackage{mdframed}
\usepackage{array} % Needed for centering in the table
\usepackage[export]{adjustbox} % loads also graphicx
\usepackage{graphicx}

\usepackage{hyperref} % Creates links in the PDF document.
\hypersetup{hidelinks} % Do not include boxes around links

\usepackage[tableposition=top]{caption}
\usepackage[vmargin={1.25in, 1.25in}, hmargin={1.5in, 1in},
            textwidth=6in,
            textheight=5.25in]{geometry}
\usepackage{microtype}
\DisableLigatures{encoding = *, family = * }

\makeatletter
\renewcommand{\@tocrmarg}{2.3em plus1fil}
\makeatother

\def\projectType{Project}
\def\graduationMonth{May}
\def\advisorName{Dr. Thomas Austin}
\def\advisorAffiliation{Department of Computer Science}
\def\firstCommitteeMember{Dr. Jon Pearce}
\def\firstCommitteeMemberAffiliation{Department of Computer Science}
\def\secondCommitteeMember{Dr. Grit Denker}
\def\secondCommitteeMemberAffiliation{SRI International}

\begin{document}

%
% ***** title
%
\title{Toward Strategy Identification and Subtask Decomposition In Task Exploration}

%
% ***** author's name
%
\author{Tom Odem}
%
% ***** graduation year
%
\copyrightyear{2025}

%
% Do not change this...
%
\ifthenelse{\equal{\projectType}{Thesis}}
           {\raggedright \parindent=30pt}
           {}

\pagestyle{empty}

%
% ***** abstract is file abs.tex
%
\abstract{abs}

%
% ***** acknowledgements in file ack.tex
%
% You can optionally have a dedication as well supported by the URI template.
%
\acknowledgements{ack}
%\dedication{ack}

\setlength{\parskip}{.1 in}
%uncomment the following for single space drafts
%\singlespace
\doublespace
\pagenumbering{arabic}

\setcounter{page}{0}

%
% ***** chapters in files chap1.tex, chap2.tex, and so on
%
\chapter{Introduction}
\label{chap:intro}

\section{Human-Machine Interaction} 

Human-machine interaction (HMI), as the name suggests, is a field that focuses on improving interactions between humans and machines. It combines methods from computer science, psychology, user experience, human factors and ergonomics, cyber security, and more to bridge the gap between humans and machines. HMI is vital for creating systems that can aid humans in completing many tasks that are present in today's technology and information centric society; whether it's a person searching the Internet on their smartphone or a surgeon performing a robot-assisted surgery, HMI research tries to make the interaction between technology and human users as seamless as possible. This idea of seamlessness is often seen as making the experience more human-like.

\section{Anticipatory Human-Machine Interaction} 

One way to make technology more human-like is for the machine system to anticipate what a user will do and act on that anticipation. Humans often anticipate each other, where one human anticipates the actions of another human's future state and acts upon it before that future state happens. An everyday example could be when a bowl of soup arrives with a spoon. The customer doesn't explicitly request a spoon, but the restaurant anticipates that it will be needed. Anticipatory HMI (aHMI) aims to introduce anticipation into human-machine systems. Instead of a spoon with a bowl of soup, it may be a suggestion to more literature upon finding something on the Internet or an automated report after a surgery. The goal of aHMI is to make the machine a proactive partner, rather than a reactive participant.

\section{Types Of Anticipation} 

In a meeting of aHMI scholars from many fields of science \cite{dagstuhl}, three integral perspectives on anticipation were identified: 1. Anticipating the intentions, beliefs, mental states, and future actions of a user 2. Anticipating the machine’s own actions 3. Anticipating the outcome of human-machine teaming. The first perspective anticipates what and how a user might complete some task. The second perspective allows the machine to anticipate what and how it will help complete the same task. The third perspective anticipates the results and consequences of the human-machine system. The hardest perspective to anticipate is perspective one. Knowing the user's current intention, beliefs, mental state, and actions is already a hard problem. Anticipating the future state of each is even harder. 

To complete the anticipation step, machines would need to engage in implicit coordination \cite{improvingTeamwork}. Right now, many systems use explicit coordination, where the computer is told exactly what to do from the click of a mouse or a key press. Implicitly realizing what is needed to complete a given task and coordinating present and future actions with that realization would make a more efficient machine partner. To aid in implicit coordination, machines should develop a theory of mind (ToM) \cite{improvingTeamwork}.

\section{Theory Of Mind} 

ToM can be described as when a being considers the state of knowledge and point of view of another being \cite{beyondNumericalRewards}. The state of knowledge is what a being knows about the task they are completing. This includes typical steps, foreseeable problems, possible solutions, and abstract ideas that together form a state of that being's knowledge. The point of view of a being is that being's interpretation and understanding of a task and its environment. In the context of aHMI, when a machine identifies what the human user knows about the current task and the user's interpretation of the solution of the task, it can start to anticipate what the human will do to complete the task.

\section{Gaps \& Challenges In aHMI}

Researchers \cite{improvingTeamwork} note three areas of aHMI that would help in adding ToM capabilities to machines: Communication, adaptability, and coordination. Gaps in communication involve challenges in the process of exchanging information with machines, machines lacking transparent processes, and an absence of a shared mental model of tasks. Adaptability sees struggles in a machine's ability to change its interface or operations in response to new or future environmental and user states. Coordination, as described earlier, sees a shortcoming in implicit coordination. 

Researchers \cite{improvingTeamwork} posit that implicit coordination requires the machine to have an organized collection of user knowledge, skills, and behaviors for a task. This organization allows the machine to understand what the user knows, what the user is able to achieve, and how the user will use their knowledge and skills to complete a task. After granting a machine with these capabilities, the next step would be for the machine to start predicting behaviors, obtain common historical knowledge, and help users in tasks.

\section{Research Objective}

Our research aims at advancing human-machine coordination by exploring how tasks are completed. The study intends to create a method that can determine key strategies for completing tasks and identify subtasks that are pertinent to completing tasks. Strategies are defined as any meaningful methods of completing a task, which may include the types of actions taken, the order of actions taken, or more. Subtasks are defined as a short sequence of actions that creates some meaningful and purposeful function. Subtasks must also be transferable, being used across different user runs, ensuring that they are useful in multiple contexts. 

Given time-series data of users completing a task in the form of user runs populated with individual actions, our method should be able to automatically identify useful strategies and subtasks. Strategies help inform the machine of user knowledge and behavior. Subtasks aim to identify user skills, knowledge, and behavior. The question this task exploration research seeks to answer is ``Given user run time-series data for a task, can useful key strategies and meaningful subtasks be automatically identified that help inform humans and machines of user knowledge, skills, and behaviors?''

The remaining sections of this paper are organized as follows. Chapter 2 reviews the literature on task exploration and builds the basis of our methodology. Chapter 3 outlines our methods. In Chapter 4, results are presented. Chapter 5 discusses the meaning of our results. Finally, Chapter 6 summarizes our main conclusions and goes over planned future work.
\chapter{Literature Review}
\label{chap:lit}

The literature review summarizes other work that is either directly tied with task exploration or is used to build the methods and requirements of a task exploration framework. There are two primary sections: The Subtask Identification section generally covers the literature that aids in identifying and defining subtasks and the Strategy Detection section focuses on what types of strategies are identified and how they are identified. There is some overlap between the two ideas within each section, but the papers are grouped by their primary purpose.

\section{Subtask Identification}

Heard et al. \cite{predictTaskPerformance} developed two Long Short Term Memory (LSTM) prediction models to associate user actions with anticipated task performance. The models use time-series data of the user's current actions and aim to predict whether the user will experience a workload conflict. A conflict occurs when a user is overloaded in multiple workload channels, such as hearing a notification while speaking. Although not immediately identifiable as a subtask identification method, the LSTMs are actually learning subtask sequences that are then associated with future states. When LSTMs learn from sequences, they essentially identify patterns that jointly occur within some target class. These joint patterns could be seen as subtasks that are linked together to predict a future state. For our research, it may be difficult to extract the identified subtasks from the LSTM in a meaningful way. Additionally, the subtasks are explaining some future state, not an overall task. The second point is more pertinent to the aim of our research, where it is important to identify the subtask structure within a task, rather than the structure that leads to some conflicted future state. Generally, as is the case in Heard et al.'s method, LSTMs are used for prediction. Our research is currently focused on exploring the tasks, not predicting them.

Heard et al. \cite{predictTaskPerformance} also introduce the concept of identifying multiple structures correlated with different views of the same user actions. It was noted that the researchers developed two LSTM models; one LSTM model is for identifying ``typical'' strategies, while the other LSTM model is for ``personal'' strategies. A typical strategy would be a strategy that is taken by many people, being typical of the recorded population. A personal strategy would be when the LSTM predicts future states based on a specific user's recorded sequences. This gives us two clear views, where a global view can be utilized for generalization and a focused view for more personalized predictions. Although our research does not look at individual task structures, this idea of different structure views can be adapted. Global structures could explain generalized processes, while local structures within the global structures could define more refined processes. This creates a hierarchical view of how tasks are completed.

A different approach to identifying subtask-like sequences would be to find macros with unique structures. A macro is a sequence of computer actions that can be turned into a single command, often used to automate some repeatable process. Typically, macros are exact copies of computer actions and corresponding parameters, but Myers et al. \cite{learningByDemo} use generalized structures and parameters to enhance macros. This allows users ``to create parameterized procedures that automate general classes of repetitive or time-consuming tasks''. Rather than defining individual macros for every possible parameter and structure correlation, a user can instead create a generalized template that can be filled in later with the exact parameters and a specific structure. 

There are two points that help with subtask identification. The first is to use macros as subtasks. According to our definition outlined in Chapter \ref{chap:intro}, subtasks must be transferable. Macros are designed to be transferable, allowing users to automate the same repeated sequence. The second insight is that subtasks are also be defined by generalized structures and parameters. By focusing on the fundamental actions in the time-series data,  useful and transferable subtasks can be found, as is done in Subchapter \ref{chap:meth}.2. If subtasks were identified on specific parameters given and rigid sequences of user actions, like typical macros, there would likely be many subtasks that fundamentally explain the same generalized meta-subtask. These meta-subasks are more transferable than a set of ungeneralized subtasks and also allow us to find the generalized structure of a given task.

Borse et al. \cite{dagstuhl} explain that one key focus to creating user interfaces that anticipate, rather than react, is a computer's ability to sense and model human attention. One of the many methods the researchers use for human attention modeling is task interleaving, where they find that the way various tasks interweave can be used to explain and anticipate a resulting attention state. Executed tasks that are not pertinent to the current meta-task could foresee diminishing attention, while tasks that are related enforce positive attention. In our research, the term task can be renamed to subtask, where interleaving subtasks help to explain the overall task that is being completed. The idea of subtask interleaving can be equated with subtask composition. Our task exploration research identifies which subtasks are useful for completing a task, how subtasks interact with each other, and how the composition of one set of subtasks differs from others. This is discussed in Subchapter \ref{chap:disc}.4 and \ref{chap:disc}.5.

Finding patterns that can be automated into macros can often use sequence mining; however, sequence mining may not be the best option for finding generalized sequences. Gervasio and Myers \cite{learningProcs} advanced the macro-generalization technique with a new method called procedure mining. In classic sequence mining, the model will identify exact sequential patterns, which does not help in finding generalized sequences. In procedure mining, the researchers compose sequences with parameterized actions, allowing the model to discover relationships between parameters to reach parameter generalization. The researchers found that classic sequence mining finds many irrelevant patterns, while their procedure mining technique can filter out many of these irrelevant patterns. This tells us two things; Our research on subtask decomposition and task exploration does not use traditional sequence mining to identify patterns as subtasks and it is able to filter out irrelevant patterns, as is done in Subchapter \ref{chap:meth}.7. Additionally, procedure mining reinforces the idea of identifying generalized sequences, rather than exact copies of actions and parameters. 

In creating a computer system that can anticipate and work with humans on a conscious level, the system requires shared intentionality. Jha and Rushby \cite{inferIntent} explain that shared intentionality, when two conscious beings understand the conscious thoughts of each other and share the same intentions, is a major requirement for human consciousness. They then posit that an extension of inverse reinforcement learning (IRL) can infer logical specifications to enable self-reflective analysis of learned information, compositional reasoning, and integration of learned knowledge. These logical specifications would then be used in a form of shared intentionality, where the system can anticipate what the user is doing and assist the user with their identified intentions. This method also enforces the idea of generalized sequences, as the IRL method generalizes exact demonstrations into logical specifications. These logical specifications can be seen as a form of subtask that identifies a generalized pattern and the locations where the pattern is typically used. Our research also attempts to identify when subtasks are typically used and, again in line with previous works, find generalized sequences over exact copies of patterns. Typical use and generalized sequences are demonstrated in Subchapter \ref{chap:disc}.5 and Subchapter \ref{chap:meth}.2, respectively.

There are some drawbacks of using an IRL method in relation to task exploration. The biggest is that IRL is a prediction technique and not an exploration technique. The technique essentially creates a new model that no longer represents the input data. Our research on task exploration aims to find how different user runs are similar or dissimilar, where as this IRL method would not be able to represent those correlations. Some identified subtasks may be extracted, but these may contain even more granular subtasks that are not used by themselves. These smaller encased subtasks would still be important to identify, as they can explain the granular interleaving of subtasks that create larger subtasks. Our method identifies subtasks that are encased within other subtasks and also focuses on exploring the correlations between individual user runs. See Subchapter \ref{chap:meth}.8 and \ref{chap:meth}.6 for implementation details.

\section{Strategy Detection}

Wang et al. \cite{subtaskAnalysis} aim to find task structures using exploratory analysis to identify subtasks and encode a sequence of actions into a sequence of subtasks. They use probability distributions to identify subtasks in simple file-selection tasks. When the probability of some set of actions decreases by a significant amount, they posit that the decrease signifies a change in the subtask being completed. After identifying subtasks, they group user runs into different strategies to find common ways that the task is completed. The grouping of user runs in order to find common strategies relates directly to the idea of having task structures. With this idea of strategies, global and local structures can be referred to as global and local strategies. 

Based on Wang et al.'s \cite{subtaskAnalysis} identification of subtasks, if there are some sets of actions that frequently happen together, they will all be grouped into a single subtask. Such a grouping leaves no room for granular subtasks that may reside within larger subtasks. For example, if there were some bi-gram subtasks (subtasks consisting of two actions) that appeared within multiple larger n-gram subtasks, they would be missed entirely and just become part of the larger subtasks. These encased subtasks are still vital to understanding how both tasks and strategies are completed, particularly with finding subtasks that are transferable across multiple strategies and tasks. Our research identifies subtasks hierarchically, where granular subtasks can be hierarchically encased within larger subtasks. Additionally, similar to Wang et al.'s work \cite{subtaskAnalysis}, our research encodes a sequence of actions into a sequence of these hierarchical subtasks. See Subchapter \ref{chap:meth}.8 for implementation details.

Borst et al. \cite{dagstuhl} present a framework for creating personalized Intelligent Tutoring Systems (ITS) using aHMI. The researchers explain that they are interested in discovering how AI-driven personalization relates to aHMI, particularly with how the AI personalizes content based on \textit{current} user state opposed to \textit{predicted} user states. Their current research aims to develop ITS for students in exploratory open-ended learning environments. Using clustering and association rule mining, the researchers identify behaviors that promote or hinder learning. Using the labels assigned from the clusters, they create supervised classifiers that can anticipate the effectiveness of the user's current learning strategy. How these researchers use association rule mining with clustering techniques is pertinent to the exploratory research our study aims to resolve. The researchers are not particularly interested in finding subtasks, but rather are finding behaviors within the action sequences that make the actions similar to each other. Applying a similar technique to our time-series data could result in clusters of unique sets of actions that are used to solve a task. 

A different IRL method from Vazquez-Chanlatte et al. \cite{learningTaskSpecs} uses non-Markovian boolean specifications to identify subtasks that can be recomposed into a larger metatask. Non-Markovian boolean specifications allow the model to express tasks with temporal dependencies to take advantage of compositionality. Although IRL was determined to be unhelpful in the pursuit of task exploration, this idea of compositionality is very important.  Runs that use a set of actions in similar compositions should be grouped. Relating back to global and local strategies, there is likely some global set of strategies that use the same set of actions. Locally, within these global strategies, there could be strategies that utilize the same set of actions in similar compositions. 

Looking at clustering implementations, Cheng and Guo \cite{activityBasedAnalysis} used factor analysis combined with hierarchical clustering to find open source software (OSS) roles. The researchers recorded counts for many important activities in OSS, like number of issues reported, number of commits made, etc. They then perform factor analysis on the count data and transform the count data into factor scores. With the user data now in a factor space, the researchers employed hierarchical clustering to produce clusters of users with similar factor scores. These clusters are then interpreted as the different roles of OSS. Although our research does not aim to identify roles, the same method can be used to identify pertinent strategies used to complete tasks. Our time-series data can be transformed into counts of actions, then follow the same methods to produce clusters as strategies. Such strategies can be seen as the aforementioned global strategies, where they are created as users who use a similar set of actions in a similar count distribution.

To introduce compositionality to a latent semantic analysis (LSA) framework, Dennis \cite{wordOrder} uses string edit distance. Traditional LSA finds meaning by the choice of words used in a document, ignoring the order of the words. While word choice is important, the order in which they appear also creates important meaning. The researcher uses string edit distance as a means to compute sentence similarity, introducing word order to LSA. String edit distance finds the ``distance'' that it takes for one sentence to be transformed into another. There are many implementations of edit distance, but the basic rules allow for adding characters, deleting characters, and substituting characters. The lower the amount of edits, the more similar two sentences are and the lower the distance. To incorporate word order in a sentence, the words can be treated as characters in a string. Similar to Dennis's \cite{wordOrder} approach, our task explorer framework can use string edit distance to add compositionality. If our time-series data can be interpreted as strings, the similarity of two user runs can be determined and compared to a larger pool of user runs. Such a method would identify local strategies within the global strategies, where users not only use the same actions in a similar count distribution but also use those actions in a similar composition.

\section{Summary}

To review, our task exploration pipeline includes the following:
\begin{itemize}
    \item Useful subtasks that are identified as macros that make up a hierarchical structure. 
    \item Subtask macros that are defined with generalized fundamental actions and with a definition process that filters out irrelevant patterns. 
    \item Two views of strategies that are identified with clustering techniques: Global strategies that cluster user runs which utilize a similar set of actions and intra-cluster local strategies which group users runs that use similar actions and have similar composition. 
    \item Global strategies that are found using factor analysis to allow the clusters to be made in a correlated factor space. 
    \item Local strategies that group user runs using string edit distance to measure similarity in composition.
\end{itemize}

\chapter{Methods}
\label{chap:meth}

This section summarizes the dataset on which the task explorer pipeline was tested, the format of input data, the vectorization techniques used on the input data, how strategies were identified, the process of defining subtasks, and the steps to encode runs with defined subtasks.

\section{Capture The Flag Dataset} 
While the task explorer pipeline can be deployed on any user run time-series data, this project uses data collected from two live capture the flag (CTF) cyber attacker events ran by SRI International. The dataset is expected to be publicly released in the fall of 2025, but a summary is included to better understand the methods and results of the developed task explorer pipeline. In each CTF event, participants were given six cyber attack tasks to complete, each with a set time limit, resulting in twelve tasks total. Every command executed by participants in the Linux environment was recorded with the time of the command's execution. The number of participants in each task varies, but there were less than 100 and more than 10 participants for all tasks.

\section{Input Data Format} 

Before deploying the task explorer pipeline, the input data must be formatted. Our data was not originally recorded as time series data, instead consisting of rows with actions and timestamps. First, these rows get turned into a single array of actions, where actions are in sequence corresponding to their time stamps. This now gives us every user's run, easily readable as the sequence of the actions they used to complete a task. An example of the time alignment process can be seen in Table \ref{dataFormatSteps} Step 1.

Next, actions are transformed into their basic action. Many actions, like commands in a command prompt, include parameters or other granular information. Our goal is to generalize many granular representations of the same actions into a small amount of corresponding fundamental actions. For example, an action for executing a python script may look like ``python python\_script.py'', but another action may look like ``python other\_python\_script.py''. These two actions are granularly different, but both have the same base action of executing a Python script. So, every action in the dataset gets turned into a fundamental action that explains the fundamental action being performed. For our particular dataset of commands being executed on a Linux command prompt, this mostly involved taking the first term of each command string. For the Python example, this would look like turning every command of the form ``python some\_python\_script.py'' into ``python''. An example of the fundamental action transformation process can be seen in Table \ref{dataFormatSteps} Step 2.

Next, actions that are fundamentally the same but are represented differently need to be identified. This can be the case when similar or exact fundamental actions are performed with different names. For example, Python can be called with the ``python'' command or the ``python3'' command. While one ensures a Python script is ran with Python 3 and the other does not, they are still performing the same fundamental action of executing a Python script. Any actions that are fundamentally the same, but are recorded as different names, should be mapped to a single useful name. An example of the identical fundamental action transformation process can be seen in Table \ref{dataFormatSteps} Step 3.

Next, ``bad'' terms are removed. Bad terms are those that do not represent any real action. In our command prompt data, this tends to be a simple spelling mistake. Any place where ``python'' is spelled as ``pthon'' or ``man'' is spelled as ``mam'' would be good examples. These terms have to be manually removed, either by creating a list of ``good'' terms or by creating a list of ``bad terms'' and pruning the rest. Our implementation uses a list of bad terms, so that the bad terms are saved for later pruning in other settings. An example of the bad term removal process can be seen in Table \ref{dataFormatSteps} Step 4.

Lastly, stopword actions are removed. These are actions that do not contribute much information to the completion of the task or that just represent a part of a different action. For example, using the ``nano'' action before executing a ``python'' action is fundamentally just executing a Python script that happened to be written before hand using ``nano''. The ``nano'' action by itself does not describe any action that adavances the completion of the task by itself. Since writing a file does not contribute to a different action other than executing a separate command, all file editing commands are marked as stopwords and are removed from all runs. This may not be the case for other tasks or datasets, so one must ensure that stopwords are carefully chosen. An example of the stop word removal process can be seen in Table \ref{dataFormatSteps} Step 5.

\begin{table} [ht]
\begin{center}
\begin{tabular}{ |l|l|l| }
    \hline
    Step \# & Input & Output\\
    \hline
    Step 1 &
    \begin{tabular}{l}
        [``nano python\_script.py'', \textbf{11:30}], \\
        {[``pthon python\_script.py'', \textbf{11:32}]}, \\
        {[``python3 python\_script.py'', \textbf{11:31}]},\\
        {[``python python\_script.py'', \textbf{11:33}]}\\
    \end{tabular} 
    & \begin{tabular}{l}
        [``nano python\_script.py'', \\
        ``python3 python\_script.py''\\
        ``pthon python\_script.py''\\
        ``python python\_script.py'']\\
    \end{tabular} \\
    \hline
    Step 2 & 
    \begin{tabular}{l}
        [``nano \textbf{python\_script.py}'', \\
        ``python3 \textbf{python\_script.py}''\\
        ``pthon \textbf{python\_script.py}''\\
        ``python \textbf{python\_script.py}'']\\
    \end{tabular} &
     \begin{tabular}{l}
        [nano, \\
        python3\\
        pthon\\
        python]\\
    \end{tabular} \\
    \hline
    Step 3 & 
    \begin{tabular}{l}
        [nano, \\
        python\textbf{3}\\
        pthon\\
        python]\\
    \end{tabular} &
    \begin{tabular}{l}
        [nano, \\
        python\\
        pthon\\
        python]\\
    \end{tabular} \\
    \hline
    Step 4 & 
    \begin{tabular}{l}
        [nano, \\
        python\\
        \textbf{pthon}\\
        python]\\
    \end{tabular} &
    \begin{tabular}{l}
        [nano, \\
        python\\
        python]\\
    \end{tabular} \\
    \hline
    Step 5 & 
    \begin{tabular}{l}
        [\textbf{nano}, \\
        python\\
        python]\\
    \end{tabular} &
    \begin{tabular}{l}
        [python\\
        python]\\
    \end{tabular} \\
    \hline
\end{tabular}
\caption{Steps To Format Input Data}
\label{dataFormatSteps}
\end{center}
\end{table}

\section{Term Frequency Vectorization} 

As described in the previous section, our data is in the form of time-series data. Each run is composed of a sequence of actions that a user takes to complete a task. Such time-series data cannot be clustered without modification. Taking inspiration from creating vectors with common OSS activity counts \cite{activityBasedAnalysis}, the aim is to create a vector of action counts. First, the total term frequency for each action is recorded over the entire set of runs for the task. 

With the total term frequency for each action, the next step is to decide the frequency \textit{n} that a term needs to be equal or greater than to be included as a term in our term frequency vector. This is done for two reasons: 1. If every possible action is included, even those that have a term frequency close to one, then there will likely be too many actions for a relatively low number of runs to perform factor analysis without getting a singular matrix. 2. Actions with a low term frequency are inherently not very important for solving the task, as they are used very infrequently. Thus, \textit{n} is manually chosen based on when there is a large dip in action term frequency and when the actions on the lower frequency counts stop serving to assist in the completion of the task. See Table \ref{tfExample}, introduced in Chapter 4, for an example with the CTF commands dataset.

With a suitable \textit{n} chosen, a list of allowed actions that have an overall term frequency greater than or equal to \textit{n} is created. Next, each run is iterated through and turned into term frequency vectors, based on the list of allowed actions. Each run is now transformed into the count of how many times every allowed action is used within the run. 

The runs are then stabilized using the square-root transformation. Stabilization ensures that high counts don't overshadow smaller, but equally important, counts. For instance, if the action ``python'' was performed 100 times in one and run only 25 times in another, the square root transformation would turn these counts into 10 and 5. This retains the fact that ``python'' was used more in one run, but also allows the counts to be more fairly compared.

To finish our term frequency vectors, each term column is stabilized. This means that zero mean and unit variance is applied to each action column. Zero mean makes each column's mean equal to zero, while unit variance means that the standard deviation of each column is equal to 1. Standardization makes these vectors ``unitless'' and also ensures that different action counts can be easily compared. An example of a term frequency vector can be found in Table \ref{factoreScoreExample} in Chapter \ref{chap:res}.

\section{Factor Analysis To Get Run Factor Score Vectors} 

Just as was done to create factors for OSS term frequency vectors \cite{activityBasedAnalysis}, factor analysis (FA) is employed to transform our term frequency vectors into factor score vectors. Before FA can be applied, a list of actions that are highly correlated needs to be gathered and the actions must be merged. This is because highly correlated terms in term frequency vectors create a singular matrix when performing FA. Singular matrices are not allowed in FA, so the options are to either remove the highly correlated terms or, in our case, merge the terms to retain their existence. Highly correlated terms are identified by looking at the correlation matrix produced from the term frequency vectors and finding terms that have correlations close to or equal to one. An example correlation matrix can be seen in Table \ref{singularMtxExample} in Chapter \ref{chap:res}. Every instance of the correlated actions gets replaced with a merged action in the original time-series data, followed by a rerun of the term frequency vectorization technique. In our CTF dataset, there are only two tasks that needed a few merged terms to complete FA. So, this is not common.

With data that will no longer cause a singular matrix, FA can be used. First, FA is run with the most number of factors possible. This gives us a number of factors equal to the number of actions within the term frequency vectors. To choose the number of factors to retain, the Kaiser criterion \cite{kaiserCriterion} is used. The Kaiser criterion selects the number of factors by retaining all factors whose eigenvalues are greater than one. This intuitively means that the retained factors explain a large amount of the data. An example of factors retained with the Kaiser criterion can be seen in Table \ref{eigenvalueExample} in Chapter \ref{chap:res}.

With the optimal number of factors found, FA is rerun on the term frequency vectors with only the optimal number of factors being produced, this time with an oblique rotation. Although highly correlated actions would be included in a single factor, oblique rotations give different factors the ability to be slightly correlated, without forcing them to be correlated \cite{obliqueRotations}. This is useful for our data, as performing one set of actions may be correlated to using another set of actions. Specifically, the Oblimin oblique rotation is used, as it is more accurate in computing factors than Promax. Promax is a faster way to approximate the Oblimin rotation \cite{obliminOblique}, but our relatively small amount of data can afford to use Oblimin.

Using the rotated factors from FA, the term frequency vectors are transformed into factor score vectors. This puts each run into the factor space, where their vector no longer represents counts but instead represents that run's placement in each factor. Every factor represents a broad activity or idea that actions can combine into. An example of a term frequency vector being transformed into a factor score vector can be seen in Table \ref{factoreScoreExample} in Chapter \ref{chap:res}.

\section{Hierarchically Clustering Runs Using Term Frequency Factor Scores} 

Hierarchical clustering is used to make clusters of runs in the factor space, as was done in OSS activity identification \cite{activityBasedAnalysis}. Hierarchical clustering allows us to cluster based on hierarchical similarities, giving clusters that can be traced to being similar or dissimilar at higher split points. Hierarchical clustering also requires the number of clusters  \textit{k} to be identified before running. To find \textit{k}, the elbow method is used with a plot of within-cluster sum of squares (WSS). For every possible \textit{k}, two through the number of runs, a new clustering instance is created for the factor score vectors. At each clustering, the WSS is computed to obtain the intra-cluster variation. This tells us how much variation between runs there typically is within each found cluster. 

Once the WSS is computed for each possible \textit{k}, the WSS measurements are plotted against the number of clusters used to compute the measurements. The elbow method is then employed to identify \textit{k}. The elbow is located where the WSS vs number of clusters plot starts to plateau, at a point where adding more clusters does not model the data any better. The identification of the elbow is automated using the the kneed python package \cite{kneed}, which implements the kneedle algorithm \cite{kneedle}. An example of this operation can be seen in Figure \ref{optimalK} in Chapter \ref{chap:res}.

After finding \textit{k}, the factor score vectors are clustered again with the optimal number of clusters \textit{k} to get the final clustering. Each cluster denotes a group of runs that is similar in the factor space. Similarity in the factor space can be based on multiple things: it can represent similar counts of different actions, the usage of a unique set of actions, a lower or higher number of a certain action, and more. These clusters are the global strategies defined in Section \ref{chap:lit}.

\section{Edit Distance Single-Linkage Hierarchical Clustering} 

Now that clusters based on action term frequency are obtained, clusters of runs that have similar composition can be identified. Intuitively, the previous term frequency clustering returns clusters of runs that generally use the same set of actions in a similar frequency. This does not include the idea of composition, where these actions are used in sequence in similar ways. Since the term frequency clusters already return groups of generally similar runs, new compositional clusters within each term frequency cluster are created. These compositional clusters are the local strategies defined in Section \ref{chap:lit}.

String edit distance is used as the similarity measurement between the composition of runs. String edit distance algorithms compare strings and do not have a direct method of comparing sequences of actions. To transform the runs into strings, a unique symbol is assigned to each action. For each action in a run, the corresponding symbol is concatenated to the corresponding string run. To compute the string edit distance between runs, the Jaro-Winkler edit distance \cite{jaroWinkler} is used. Defined in Equation \ref{jaro-winkler}, |$s_i$| is the length of string $s_i$, $m$ is the number of matching characters, and $t$ is the number of transpositions. Unlike typical string edit distance metrics, Jaro-Winkler computes distance using a similarity score between two strings that is normalized between zero and one. This is then turned into a distance metric by negating the score from one. A distance of zero means the strings are the same, while a distance of one means that strings are completely different. This normalization helps us directly compare edit distance scores across different pairs of strings. Another major way that Jaro-Winkler differs from other string edit distance metrics is that it considers the characters that already exists but just needs to be moved. Most string edit distance metrics treat these out of order characters as non-matching characters that need to be deleted and inserted again, rather than simply moved. Additionally, the Jaro-Winkler distance places more emphasis on the beginning of strings. So, in our case, runs that start out using the same actions will be closer than runs that converge later. This is important as it utilizes the idea of task interleaving. An example of this can be seen in Table \ref{cluster6CompositionalClusters} in Chapter \ref{chap:res}.

\begin{equation}
sim_j = \frac{1}{3}(\frac{m}{|s_1|}+\frac{m}{|s_2|}+\frac{m-t}{m})
\label{jaro-winkler}
\end{equation}

For each term frequency cluster, the runs that are present within the grouping are gathered. These runs are not term frequency vectors, but are the original runs as time-series data. DBSCAN is then used to cluster these runs based on the Jaro-Winkler distance. The minimum points in DBSCAN is set to two, which allows clusters of any size to be created. This is imperative, especially with the reduced number of runs on which to cluster. Two clusters should be able to form a cluster if they are very similar even if there are no other clusters similar to them. Interestingly, when setting DBSCAN's minimum points parameter to two, the clustering algorithm turns into single-linkage hierarchical clustering. Despite this, DBSCAN is still used from Scikit-learn \cite{scikit-learn} due to easily being able to define custom distance metrics.

To find DBSCAN's maximum distance parameter $\epsilon$, the elbow/knee method is used again. Nearest neighbor clustering is used to obtain the largest distance out of the two nearest neighbors of every point. These distances are plotted from highest to lowest and the kneed algorithm \cite{kneed} is used to automatically find the elbow or knee in the plot, as seen in Figure \ref{elbowCompositional60} in Chapter \ref{chap:res}. The intuition of this is to find a maximum distance that will not be high enough to form large and useless clusters, but also not be too low to never form relevant clusters. If no elbow or knee is found, $\epsilon$ is set to the median of the distances. In the case where there are only two runs in a term frequency cluster, $\epsilon$ is set to 0.2. Throughout all of the clusterings, $\epsilon$ values near 0.2 appeared frequently. Additionally, a distance of 0.2 means that two runs have a similarity of at least 0.8, which requires two runs to have a very similar composition to be clustered together. Importantly, clusters grouped within a ``-1'' cluster are not similar. A ``-1'' cluster designates that the grouped runs are not close enough to any other run to be a part of a cluster.

\section{Defining Subtasks} 

Subtasks are defined separately from strategy clustering. When defining subtasks, an optional dictionary of previously defined subtasks is accepted. This allows cross-compatibility between different tasks and ensures previously defined subtasks are not redefined. To define new subtasks, all runs from a task, whether they were used in clustering or not, that have one or more actions are gathered. The runs are then turned into a single document, where each run is concatenated to the last. Repeated actions are turned into single instances. For example, ``python python python python'' would be turned into ``python''. This is because the definition of subtasks does not care about a single action being taken over and over again. Subtasks focus on the interactions between different actions and how one action may incite another.

As Chapter \ref{chap:intro} defines them, a subtask is a unique and transferable sequence of actions that gives some meaningful insight. To gain meaningful insight, NLTK \cite{NLTK} is employed  to find collocations with n-gram sizes of two, three, and four. Collocations are n-grams that are present with higher frequency than chance, like the tri-gram ``United States of America'' in English. N-grams are limited to a maximum size of four for two reasons: 1. NLTK does not support collocations of larger sizes and 2. The CTF dataset produces very few quad-grams per task, indicating that penta-grams would be even rarer and likely not worth the effort of manual implementation. 

There are many n-grams found in our dataset that are not very relevant. To further refine the n-grams into collocations, pointwise mutual information (PMI) is used to select the n-grams that occur at greater probability than if the singular actions were independent. Defined in Equation \ref{PMI}, $a$ is an action and $b$ is another neighboring action, $P(a,b)$ is the probability that the two actions occur together, and $P(a/b)$ is the probability that $a$ or $b$ occur independently. This equation can be easily translated to n-grams of any size by adding more action variables. Where n-grams are sequences of actions that occur in the document somewhere, PMI ensures that the n-gram has a high probability of being used purposefully. This is what turns an n-gram into a collocation. A PMI greater than zero indicates a probability greater than chance, so n-grams that have a PMI greater than zero are retained as collocations.

\begin{equation}
PMI(a,b) = log(\frac{P(a,b)}{P(a)P(b)})
\label{PMI}
\end{equation}

As the definition states, subtasks must be transferable. To ensure that defined subtasks are transferable, collocations only get turned into subtasks if they occur in more than a single run. An example of the number of n-grams, collocations, and subtasks found through this process can be found in Table \ref{ngramCollocationSubtask} in Chapter \ref{chap:res}. After the subtasks are found, each is given a unique name. The name is in the form of ``st[n-gram size][unique identifier]'', where st stands for subtask, n-gram size is the number of actions within the subtask, and the unique identifier is an integer that only that sequence of actions is assigned within the n-gram size. A subtask of the form ``python, hydra'' may be named ``st21'', which would mean it is a subtask with two actions and has a unique id of 1. When given a dictionary of previously defined subtasks, unique identifiers are started at the current highest unique identifier for the n-gram size plus one. Some examples of subtasks and their names can be found in Table \ref{foundSubtasks} in Chapter \ref{chap:res}.

\section{Hierarchically Encoding Runs Using Subtasks}

After defining subtasks, they need to be related to individual runs. This is done using a hierarchical method of encoding, creating a top-level encoding of the run and storing the hierarchy of all subtasks within the run. For each run within a task, a three step encoding process is used. First, all subtasks of any size are identified within the run. When identifying subtasks, the starting and ending indices of the subtask are recorded. 

Second, ``encased'' subtasks are found. An encased subtask is a subtask that exists within another subtask in the run. If ``python hydra python'' were a tri-gram subtask, a ``python hydra'' bi-gram subtask would be encased within it. The process starts from lower n-gram sizes first, then increments to the next size, to ensure that subtasks are encased sequentially based on n-gram size. So, st21 may be encased by st31, which may be encased by st41, rather than st21 and st31 being both encased by st41.

After encasing all subtasks, the third step is to identify all of the overlaps between the remaining top-level subtasks and update their leads. One subtask “leads” into another subtask if the first subtask starts before the second subtask, but is completed within the second subtask. Overlaps are found when the starting indice of one top-level subtask occurs before the ending indice of another. After this final step, the run is encoded.

There are two methods to represent an encoded run. The first is seen in Figure \ref{hierarchicalDiagrams} in Chapter \ref{chap:res}, where every subtask is shown in a diagram as lines at various height levels and colors. Each height level and color represents a different n-gram size for the subtasks. The length of the line is equal to the length of the n-gram and the beginning of a subtask's line is equal to its starting index in the run. One can identify encased subtasks where a higher subtask line covers a subtask line below it. Leads can be found by seeing where the subtask lines overlap.

The second way an encoded run is represented is by a string version of the top-level encoding, exemplified in Table \ref{stringEncoding} in Chapter \ref{chap:res}. Every subtask and action that is not encased gets concatenated in sequence according to their starting index, with a space in between each term. When a subtask leads into another, a right arrow ``->'' is printed connecting them. The right arrow indicates that the subtask on the left leads into the proceeding subtask on the right.

\chapter{Results}
\label{chap:res}

This section looks at an example of results produced from the task explorer pipeline. Unless otherwise noted, this section focuses on a singular example task. The task of the users was to gain access to a remote server to find a file within the server. The presented results, while mainly depicting a single task, are representative of all twelve tasks the task explorer pipeline was run on.

\section{Term Frequency Threshold Example}

\begin{table}
\begin{center}
\begin{tabular}{ |l r| }
    \hline
    \multicolumn{2}{|c|}{Included} \\
    \hline
        python & 221 \\
        hydra & 146 \\
        grep & 73 \\
        man & 31 \\
        cut & 16 \\
        ffuf & 12 \\
        mv & 8 \\
        dirb & 7 \\
        nmap & 6 \\
        gzip & 5 \\
        setxkbmap & 5 \\
        curl & 4 \\
        strings & 4 \\
        exploitdb & 4 \\
     \hline
     \multicolumn{2}{|c|}{Not Included}\\
     \hline
         cp & 3 \\
         head & 3 \\
         pwd & 3 \\
         mkdir & 2 \\
         wc & 2 \\
         chmod & 2 \\  
         ... & ...\\
     \hline
\end{tabular}
\caption{Example Actions And Total Task Term Frequency Count With Term Frequency Cut-Off.}
\label{tfExample}
\end{center}
\end{table}

Table \ref{tfExample} shows an example of the gathered total task term frequency for the example task. The term frequency threshold \textit{n} was chosen as four for this task. This threshold was chosen due to the term frequency counts starting to dip into a plateau and because the actions in lower frequency counts were not as important to the context of finishing the task. In this task, ``exploitdb'' would be a vital tool when used appropriately, where as ``cp'', ``head'', and ``pwd'' could be substituted by other actions or used in many non-essential ways. Thus, the threshold of four was chosen to include the important actions and disregard the less helpful ones.

\section{Factor Analysis Example} 

\begin{table}
\begin{center}
\begin{tabular}{ |c|c|c|c|c|c|c||c|c|c|c|c|c|c|c|}
    \hline
    Factor & 1 & 2 & 3 & 4 & 5 & 6 & 7 & 8 & 9 & 10 & ... \\
    \hline
        Eigenvalue & 2.27 & 2.0 & 1.69 & 1.46 & 1.22 & 1.1 & 0.9 & 0.87 & 0.85 & 0.55 & ... \\
     \hline
\end{tabular}
\caption{Example Factors And Their Associated Eigenvalues With A Kaiser Criterion Cut Off At Factor 6}
\label{eigenvalueExample}
\end{center}
\end{table}

Table \ref{eigenvalueExample} shows the eigenvalues for ten of fourteen factors computed for this example task. According to the Kaiser criterion, only factors 1 through 6 were retained. Factors 7 through 14 had eigenvalues less than one, indicating that they should not be selected.

\begin{table}
\begin{center}
\begin{tabular}{ |l|r| }
    \hline
    \multicolumn{2}{|c|}{Term Frequency Vector} \\
    \hline
        curl&       -0.27 \\
        cut &        -0.20 \\
        dirb &       -0.26 \\
        exploitdb&   -0.15 \\
        ffuf      &  -0.25 \\
        grep       & -0.49 \\
        gzip      & -0.15 \\
        hydra     &   0.72 \\
        man        &  2.43 \\
        mv         & -0.31 \\
        nmap       & -0.27 \\
        python     & -1.44 \\
        setxkbmap  & -0.21 \\
        strings    & -0.21 \\
         \hline
\end{tabular}
\quad
\begin{tabular}{ |l|r| }
    \hline
    \multicolumn{2}{|c|}{Factor Score Vector} \\
    \hline
        1 & -0.08091022\\
        2 & -0.46977517\\
        3 & -0.40043612\\
        4 & -0.31993403\\
        5 & -0.2624389\\
        6 & 1.25976309\\
         \hline
\end{tabular}
\caption{Example Term Frequency Vector And Its Corresponding Factor Score Vector}
\label{factoreScoreExample}
\end{center}
\end{table}

Table \ref{factoreScoreExample} shows the term frequency vector for a single user in the example task and its corresponding factor score vector. The term frequencies for each allowed action were transformed into the six dimensional factor space. Each factor score represents the term frequency vector's placement in each factor.

\begin{table}
\begin{center}
\begin{tabular}{ |c|c|c|c|c|c|c| }
    \hline
    &curl&dirb&ffuf&hydra&nmap&sqlmap\\
    \hline
    curl&NaN&-0.24&0.31&-0.46&-0.24&-0.24\\
dirb&-0.24&NaN&-0.19&-0.24&-0.13&\textbf{\underline{1.0}}\\
ffuf&0.31&-0.19&NaN&-0.36&-0.19&-0.19\\
hydra&-0.46&-0.24&-0.36&NaN&-0.24&-0.24\\
nmap&-0.23&-0.12&-0.19&-0.24&NaN&-0.13\\
sqlmap&-0.24&\textbf{\underline{1.0}}&-0.19&-0.24&-0.13&NaN\\
         \hline
\end{tabular}
\caption{Example Of A Correlation Matrix That Would Result In A singular Matrix During FA}
\label{singularMtxExample}
\end{center}
\end{table}

This example task did not have any highly correlated actions that resulted in a singular matrix. However, a different task that did have highly correlated actions will be briefly covered. Table \ref{singularMtxExample} shows this other task's correlation matrix, which would result in a singular matrix. The correlation matrix shows that ``sqlmap'' and ``dirb'' have a correlation of 1.0, which means that they are both used at the same frequency in the runs they exist in. To allow factor analysis to continue, every instance of ``sqlmap'' and ``dirb'' is transformed into ``sqlmapdirb''.

\section{Formed Term Frequency Clusters} 

\begin{figure}[h]
\begin{center}
\includegraphics[width=9cm]{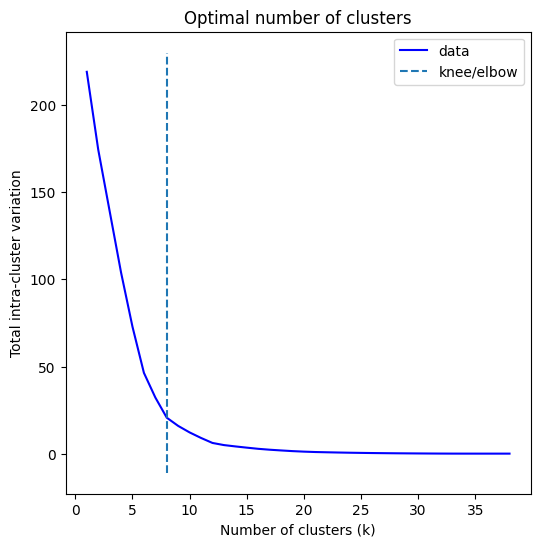}
\caption{Example Task's \textit{k} Found Via The Kneedle Algorithm}
\label{optimalK}
\end{center}
\end{figure}

The WSS computations and optimal number of clusters \textit{k} found from the automated kneedle algorithm can be seen in Figure \ref{optimalK}. Visually, the elbow lies somewhere between 5 and 10 and the Kneedle algorithm identifies 8 as the optimal \textit{k}. 

\begin{figure}[h]
\begin{center}
\includegraphics[width=12cm]{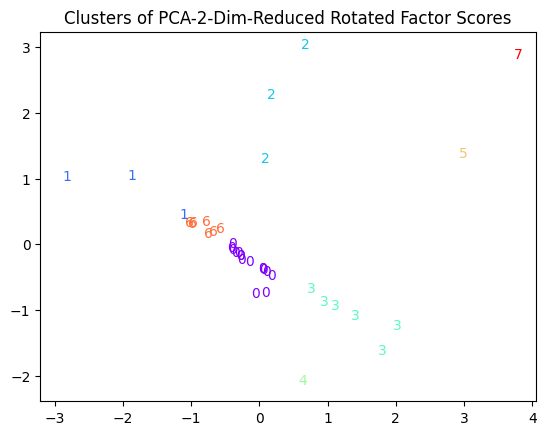}
\caption{Example Task's Two Dimensional Representation Of \textit{k} = 8 Clusters}
\label{2DimClusters}
\end{center}
\end{figure}

After clustering with \textit{k} = 8, clusters of runs that are separated hierarchically within the factor space are obtained. Figure \ref{2DimClusters} shows a two dimensional representation of the clusters. Principal component analysis was used to transform the factor space's six dimensions into two dimensions. The two axes are the two retained principal components that best represent the six dimensional factor data. Each number represents the cluster that a run was grouped into, plotted according to the new two dimensional space.

\begin{figure}[h]
\begin{center}
\includegraphics[width=10cm]{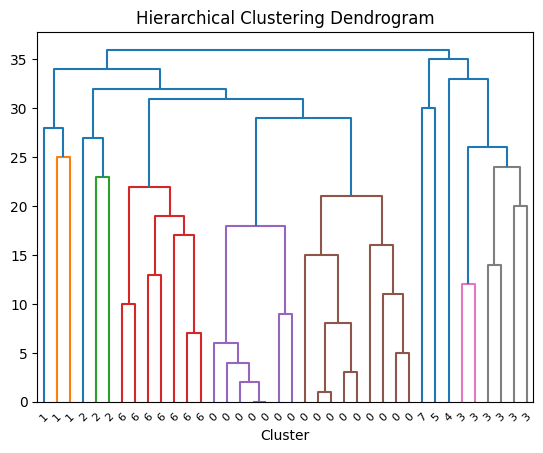}
\caption{Dendrogram Produced From Hiearchical Clusters In Example Task}
\label{dendrogramExampleTask}
\end{center}
\end{figure}

The hierarchical structure of the example task's clusters can be seen in Figure \ref{dendrogramExampleTask}. Each cluster denotes a group of runs that are similar to each other in the factor space, with each factor representing some broad activity. Most runs are in clusters with other runs, with just clusters 4, 5, and 7 having a single run each. When tracing clusters farther up the hierarchical structure, it can be determined where clusters diverge. 

Although finding exactly how these clusters diverge in the dendrogram would mean interpreting every action's influence on each factor, the run frequency of each action within every cluster can be used to determine a likely point of splitting. Table \ref{clusterActions} displays the run frequency of each action per cluster for the actions that exist within the clustered runs. Assumptions can then be made on split points using these percentages and the dendrogram. For example, clusters 1 and 6 are split at one level above their clusters. Both have heavy usage of hydra, python, and man. However, cluster 1 groups runs that all use grep, while cluster 6 groups runs that all use hydra. The usage of grep or hydra likely has a high influence in the factor space, causing these runs to be split into different groups.

\begin{table}
\begin{center}
\begin{tabular}{ |l|r| }
    \hline
    \multicolumn{2}{|c|}{Cluster 0} \\
    \hline
    python & 85.71\%\\
    hydra & 52.38\%\\
    man & 23.81\%\\
    curl & 4.76\%\\
    setxkbmap & 4.76\%\\
    \hline
\end{tabular}
\quad
\begin{tabular}{ |l|r| }
    \hline
    \multicolumn{2}{|c|}{Cluster 1} \\
    \hline
    grep & 100.00\%\\
    python & 66.67\%\\
    hydra & 33.33\%\\
    man & 33.33\%\\
    curl & 16.67\%\\
    \hline
\end{tabular}
\quad
\begin{tabular}{ |l|r| }
    \hline
    \multicolumn{2}{|c|}{Cluster 2} \\
    \hline
    python & 100.00\% \\
    hydra & 100.00\%\\
    man & 33.33\%\\
    ffuf & 33.33\%\\
    setxkbmap & 33.33\%\\
    \hline
\end{tabular}

\begin{tabular}{ |l|r| }
    \hline
    \multicolumn{2}{|c|}{Cluster 3} \\
    \hline
    dirb & 100.00\%\\
    hydra & 100.00\%\\
    python & 66.67\%\\
    ffuf & 33.33\%\\
    man & 33.33\%\\
    \hline
\end{tabular}
\quad
\begin{tabular}{ |l|r| }
    \hline
    \multicolumn{2}{|c|}{Cluster 4} \\
    \hline
    python & 100.00\%\\
    strings & 100.00\%\\
    nmap & 100.00\%\\
    grep & 100.00\%\\
    hydra & 100.00\%\\
    \hline
\end{tabular}
\quad
\begin{tabular}{ |l|r| }
    \hline
    \multicolumn{2}{|c|}{Cluster 5} \\
    \hline
    python & 100.00\%\\
    grep & 100.00\%\\
    exploitdb & 100.00\%\\
    man & 100.00\%\\
    wfuzz & 100.00\%\\
    ffuf & 100.00\%\\
    \hline
\end{tabular}

\begin{tabular}{ |l|r| }
    \hline
    \multicolumn{2}{|c|}{Cluster 6} \\
    \hline
    hydra & 100.00\%\\
    man & 66.67\%\\
    python & 66.67\%\\
    nmap & 33.33\%\\
    gzip & 16.67\%\\
    \hline
\end{tabular}
\quad
\begin{tabular}{ |l|r| }
    \hline
    \multicolumn{2}{|c|}{Cluster 7} \\
    \hline
    curl & 100.00\%\\
    hydra & 100.00\%\\
    python & 100.00\%\\
    grep & 100.00\%\\
    strings & 100.00\%\\
    \hline
\end{tabular}
\caption{The Percentage Of Runs In Each Cluster That Include The Actions That Exist Within Each Cluster}
\label{clusterActions}
\end{center}
\end{table}

\section{Formed Compositional Clusters} 

Each term frequency cluster goes through the compositional clustering algorithm, making clusters within the term frequency clusters that represent similar run composition. Table \ref{cluster6CompositionalClusters} shows the compositional clustering results for the term frequency cluster 6. Two runs were close enough to be clustered together, whereas the remaining four runs were not close enough to be clustered with any other run. Runs that are clustered together have similar composition and also share the same term frequency cluster. Looking at the two runs in compositional cluster 0, one only contains repeated uses of hydra and the other is filled primarily with repeated uses of hydra. From a Jaro-Winkler distance perspective, this is a very short distance to match the strings. The four clusters in cluster -1 are different enough to not fall within the used $\epsilon$.

\begin{table}
\begin{center}
\begin{tabular}{ |l|}
    \hline
    \multicolumn{1}{|c|}{Composition Cluster 0} \\
    \hline
        hydra
        hydra
        hydra
        hydra
        hydra
        hydra
        hydra
        ...\\
        hydra
        hydra
        hydra
        man
        man
        hydra
        hydra
        hydra
        ...
        python\\
    \hline
    \multicolumn{1}{|c|}{Compositional Cluster -1} \\
    \hline
        hydra man man hydra man man hydra man hydra hydra\\
        hydra hydra man hydra gzip gzip ... hydra hydra ... python\\
        man python ... hydra man man hydra ... python ping .sh nmap -\\ - nmap python man hydra ...\\
        python python nmap nmap hydra hydra ... nmap hydra hydra hydra\\
    \hline
\end{tabular}
\caption{Compositional Clusters Found In Term Frequency Cluster 6. Note: -1 Means Runs Were Not Close Enough To Cluster}
\label{cluster6CompositionalClusters}
\end{center}
\end{table}

\begin{figure}[h]
\begin{center}
\includegraphics[width=8cm]{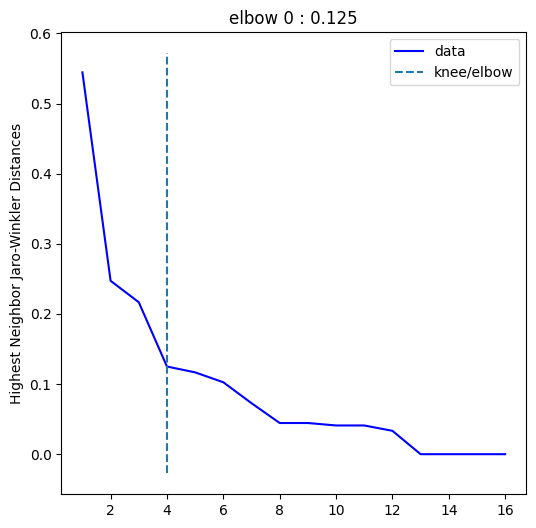}
\caption{The $\epsilon$ Elbow Found Via The Kneedle Algorithm For Compositional Clustering In Term Frequency Cluster 0}
\label{elbowCompositional60}
\end{center}
\end{figure}

Figure \ref{elbowCompositional60} shows an example $\epsilon$ elbow found using the kneedle algorithm for the term frequency cluster 0, as term frequency cluster 6 had no elbow or knee and used the median distance. The maximum neighbor distances are plotted from highest to lowest on the y axis, with the x axis simply being increments of 1. The best $\epsilon$ was found to be 0.125, which was then used as the maximum Jaro-Winkler distance with which two runs could be clustered.
 
\section{Defined Subtasks} 

For the example task, many of the initial found n-grams were disregarded to make a smaller set of relevant subtasks. Table \ref{ngramCollocationSubtask} displays the number of n-grams found, the number of corresponding collocations found, and the corresponding final number of subtasks identified for each n-gram size. Creating collocations from the n-grams using PMI did not change the total number of candidate subtasks by much, but only selecting collocations that are transferable greatly refined the final number of subtasks. Transferability ensures that multiple user runs utilized the subtask, while PMI ensures that a subtask is probabilistically purposeful and not randomly executed.

\begin{table}
\begin{center}
\begin{tabular}{ |c|c|c|c|}
    \hline
        n-gram size & \# n-grams & \# collocations & \# subtasks \\
    \hline
        4 & 164 & 164 & 4 \\
    \hline
        3 & 96 & 91 & 11 \\
    \hline
        2 & 142 & 136 & 16 \\
    \hline
\end{tabular}
\caption{The Number Of N-grams, Collocations, and Subtasks Found In The Example Task}
\label{ngramCollocationSubtask}
\end{center}
\end{table}
\begin{table}
\begin{center}
\begin{tabular}{ |l|r|}
    \hline
    \multicolumn{2}{|c|}{quad-gram subtasks} \\
    \hline
        man python hydra man & st42 \\
        hydra man hydra python & st43 \\
    \hline
    \multicolumn{2}{|c|}{tri-gram subtasks} \\
    \hline
        hydra man hydra & st30 \\
        man hydra python & st310 \\
    \hline
    \multicolumn{2}{|c|}{bi-gram subtasks} \\
    \hline
        hydra man & st20 \\
        hydra python & st24 \\
    \hline
    
\end{tabular}
\caption{Examples Of Subtasks Found In Example Task}
\label{foundSubtasks}
\end{center}
\end{table}

\section{Hierarchical Encoding} 

\begin{figure}[h]

\begin{subfigure}[h]{0.5\textwidth}
\fbox{\includegraphics[width=6cm]{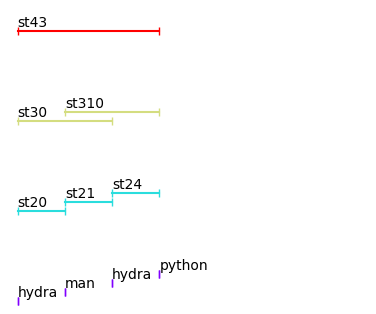}}
\caption{Run In Comp. Clust. 0}
\end{subfigure}
\quad
\begin{subfigure}[h]{0.5\textwidth}
\fbox{\includegraphics[width=6cm]{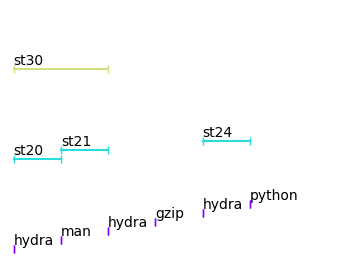}}
\caption{Run In Comp. Clust. -1}
\end{subfigure}

\begin{subfigure}[h]{0.5\textwidth}
\fbox{\includegraphics[width=9cm]{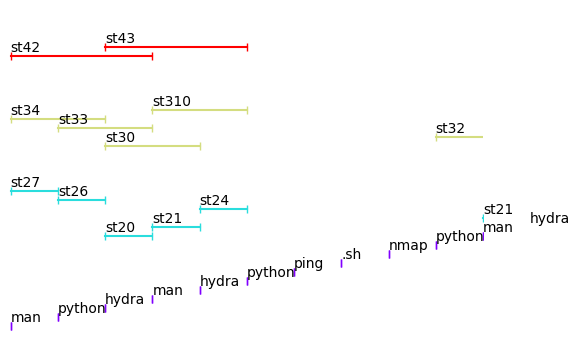}}
\caption{Run In Comp. Clust. -1}
\end{subfigure}
\caption{Hierarchical Encoding Diagrams For Some Runs In Compositional Clusters From Term Frequency Cluster 6 }
\label{hierarchicalDiagrams}
\end{figure}

A few hierarchical encoding diagrams from the example runs in Table \ref{cluster6CompositionalClusters} can be seen in Figure \ref{hierarchicalDiagrams}. Looking at Figure \ref{hierarchicalDiagrams}.a, the original run at the bottom level can be seen in purple. The run is shortened because repeat actions are removed. At the next level, the blue bi-gram subtasks found in the run are plotted, followed by the yellow tri-gram subtasks, and finally the red quad-gram subtask at the highest level. The quad-gram subtask st43 encases st30 and st310, as it spans both lengths. The subtask st30 encases st20 and st21, while st310 also encases st21 as well as st24. The same visual method can be applied to the remaning diagrams to identify the hierarchical subtask structure of the runs.

%942 a
%477 b
%813 c

For the same runs viewed as hierarchically encoded diagrams, their top-level string encoding can be viewed in Table \ref{stringEncoding}. Instead of reading the runs as long sequences of individual actions, they can now be read as short sequences of subtasks and actions. The string encoding for \ref{stringEncoding}.a is simply st43, as all actions and subtasks are encased within the hierarchical subtask st43. Encoding \ref{stringEncoding}.b has st30, encasing other subtasks, complete before executing the gzip action followed by the st24 subtask. Encoding \ref{stringEncoding}.c starts with st42, which leads into and completes in st43 before continuing onto ping, .sh, nmap, and st32.

\begin{table}
\begin{center}
\begin{tabular}{ |l|}
    \hline
    \multicolumn{1}{|c|}{Composition Cluster 0} \\
    \hline
        (a)\\
        hydra
        hydra
        hydra
        man
        man
        hydra
        hydra
        hydra
        ...
        python\\
        \textbf{st43}\\
    \hline
    \multicolumn{1}{|c|}{Compositional Cluster -1} \\
    \hline       
        (b)\\
        hydra hydra man hydra gzip gzip ... hydra hydra ... python\\
        \textbf{st30 gzip st24}\\
        \hline
        (c)\\
        man python ... hydra man man hydra ... python ping .sh nmap\\ nmap python man hydra ...\\
        \textbf{st42 -> st43 ping .sh nmap st32}\\
    \hline
\end{tabular}
\caption{Top-Level Hierarchical String Encodings For Some Runs In Compositional Clusters From Term Frequency Cluster 6.}
\label{stringEncoding}
\end{center}
\end{table}

\chapter{Discussion}
\label{chap:disc}

This section includes an interpretation of the results from Chapter \ref{chap:res} and the definitions of numerous useful statistics that are computed with the results. Additionally, a Task Explorer graphical user interface (GUI) built for easy visualization of the results is presented. It is also pertinent to remember that the results are selected from a single task, but are representative of what is produced from other tasks within the dataset.

\section{Formed Term Frequency Clusters} 

Figure \ref{2DimClusters} shows nicely separated clusters. There is no overlap between clusters, meaning that useful clusters are being formed from the automated task explorer pipeline. The most right run within cluster 1 is close to the runs of cluster 6, but it is important to remember that the plot is a two dimensional representation of the six dimensional factor space the example task was clustered on. While in two dimensions the run is close to a different cluster, in six dimensional factor space it is likely much farther away and closer to the other runs in cluster 1.

It can further be observed that these groupings are being split at reasonable places by looking at the dendrogram in Figure \ref{dendrogramExampleTask} and referencing Table \ref{clusterActions}. As described in Chapter \ref{chap:res}, assumptions can be made about how clusters are formed by looking at the run frequency of every action in each cluster. Compositional cluster 2, 3, and 6, for example, all have runs that use hydra 100\% of the time. However, they differ by the fact that cluster 2 also has a 100\% usage of python, cluster 3 having a 100\% usage of dirb, and cluster 6 having no other 100\% usage. 

While looking at this breakdown is helpful to understand what these clusters are made of, it is important to remember that the clusters are formed from factor scores derived from term frequency counts. So, the clusters do not perfectly mirror subsets of runs that have 100\% of one set of actions compared to another cluster. An example can be seen by viewing compositional cluster 0 in Table \ref{clusterActions}. There is no action that exists 100\% of the time, which is a result of clustering in the factor space. Although there is no clear connections between these runs from looking at pure action run frequency, the factor space allows the runs to be correlated from their use of actions, frequency of those actions, and the actions they may not have used. 

This is also seen when viewing singular run clusters like compositional clusters 4, 5, and 7 in Table \ref{clusterActions}. These clusters formed around only single runs, despite their set of actions being fairly similar to other clusters. The factor space is the differentiating factor. While these clusters may use a similar set of tools, they likely use them in different frequency, maybe using rarer actions more often and frequent actions less frequently. The clusters are formed from what the computed factors found meaningful about the given action distributions.

Despite the clusters not forming perfect sets of actions, they are generally creating what can be interpreted as unique sets of actions that can be viewed as strategies that users take to complete the task. Due to this, and for ease of understanding by those who do not want to delve into the factor space, all found compositional clusters were dubbed as \textit{Bag of Tools (BoT)} Strategies. BoT strategies are the realization of the global strategies extensively discussed in the Literature Review section. Runs inside of BoT strategies are using similar sets of actions in similar count distributions. These strategies represent a global view of sets of actions and action count distributions that users employ to complete a given task.

BoT strategies help to inform us of user knowledge and behavior. Knowledge can be extracted from the set of tools the user is using to complete the task, telling us the tools that a set of users is knowledgeable about. The BoT strategy that a user run is located within identifies the general behavior of the user, indicating a high use of the actions found in the strategy.

\section{Formed Compositional Clusters} 

Accordingly, compositional clusters formed within BoT strategies become the local strategies discussed and defined in the Literature Review. These are clustered runs that are not only using a similar set of actions, but are also using them in a similar way compositionally.  Compositional clusters are named as \textit{Echo} strategies, where users are ``echoing'' what others are doing. So, every BoT strategy can have local Echo strategies that group similar compositions of actions within the BoT strategy.  

Three out of the eight BoT strategies identified in the example task had Echo strategies within them. Other tasks are similar, with only about half of BoT strategies containing any Echo strategies. This tells us that many of these BoT strategies do not have a unique composition to solve the task, given a set of similar actions and action count distributions. This non-uniformity can be seen in Table \ref{cluster6CompositionalClusters}, where only two runs were grouped into echo strategy 0 within BoT strategy 6. BoT strategy 6 was clustered with all runs having hydra, where echo strategy 0 groups the two runs that are primarily a repeating use of hydra. The other runs, which are in the non-strategy cluster of -1, all use hydra far less and in combination with a more diverse and frequent amount of other actions. Additionally, none of the non-grouped runs are compositionally close to each other and are thus not clustered into an echo strategy. This shows that our string edit distance clustering technique is accurately grouping together runs that are compositionally similar, while excluding runs that do not have similar composition to others. If a BoT strategy contains echo strategies, then there is a clear way to use the actions given to complete the task. Consequently, if no echo strategies are present, there is no clear composition in which the given actions are used to complete the task.

Echo strategies help to inform us of user knowledge and behavior. Knowledge is identified from the composition of a user's actions, which informs us if they are knowledgeable on how to utilize the actions. Echo strategies detect behavior through similar run composition. User runs that have similar composition also have similar behavior.

\section{Defined Subtasks} 

 In agreement with our definition of a subtask in Chapter \ref{chap:intro} and in completion of our research question, transferable and useful subtasks were identified. Looking at Table \ref{ngramCollocationSubtask}, the subtask identification system within the task explorer pipeline was able to refine a large number of n-grams into a useful set of transferable subtasks. All retained n-grams, becoming a subset of collocations, are probabilistically being used purposefully. The subset of collocations then get turned into subtasks, ensuring transferability. By being transferable and having a probable purpose, the identified subtasks are useful to the completion of a task.

 As observed in Table \ref{foundSubtasks}, the identified subtasks could be viewed as macros. However, these are not typical macros. They are hierarchical macros that are built from smaller units of activities. Each example bi-gram subtask is an activity that takes place in the tri-gram examples. The tri-gram subtasks are activities that take place within the quad-gram examples. These are not sequences that simply complete one activity; they are building blocks that form together to create a sequence of activities to complete some macro-action.

 An interesting observation, as demonstrated in Table \ref{ngramCollocationSubtask}, is that the number of subtasks identified per n-gram size decreases as the n-gram size increases. One might expect that when there are so many smaller subtasks to build larger subtasks, there would be many more combinations of large subtasks than smaller ones. Instead, the number of subtasks seem to considerably shrink at each n-gram size. This tells us that while there are many combinations of smaller subtasks, there are purposeful and established sequences that complete macro-actions. These sequences are not random combinations of smaller subtasks, indicating that the composition of actions and subtasks does matter.

\section{Hierarchical Encoding} 

Only identifying subtasks is not helpful for exploring a task, which is why the task explorer pipeline encodes and visualizes each run hierarchically. Figure \ref{hierarchicalDiagrams} visualizes three runs in BoT strategy 6 as hierarchical encoding diagrams. Each diagram allows a viewer to visualize all the subtasks within a run and how the subtasks are composed together to create a sequence of subtasks and actions that completes the task. This is helpful both for understanding the hiearchical structure of subtasks in a run and to identify how subtasks interact with each other.

The top-level hierarchical string encodings in Table \ref{stringEncoding} allow us to interpret runs as the interactions between the top-level subtasks and actions. Instead of seeing every subtask within the run, only the top-level ones are displayed. The overlaps between subtasks are seen with the right arrow ``->'', indicating when one subtask is leading into and completing within the proceeding subtask.

This top-level encoding also allows viewers to quickly see if the user is doing something typical or not. If the majority of a top-level string encoding is filled with high n-gram subtasks, then the user is completing activities that other users also completed. In contrast, if the encoding has very few or just small n-gram subtasks, then the user was not following an established sequence to complete the task. A good mix of subtasks, actions, and n-gram sizes shows that a user is following a well established routine while adapting to their current task.

Subtasks, and their corresponding hierarchical encodings, can be used to suggest user skill level. A user run that is encompassed by larger subtasks shows that the user is following a well established routine that their peers also follow. If we assume that the users who are completing a task are skilled, then we can say that a user who is following their peers is also skilled. Alternatively, if a user has few or no subtasks, then the user is not following what their skilled peers have done and can be assumed to have a low skill level.

Subtasks also help to inform us of user knowledge and behavior. If a user run has many subtasks, then it can be assumed that they are knowledgeable of the actions they utilize, as their skilled peers also used the same actions in the same way. Subtasks identify general user behavior through short subsequences of actions that are found across user runs.

\section{Interesting Statistics} 

With the information gained from identifying BoT strategies, echo strategies, and subtasks, there are many interesting statistics that can be computed to gain meaningful insight into how tasks are completed. There are many more interesting and helpful statistics that can be computed with the task explorer pipeline's results, but those listed are the statistics that were most prominently needed in our current research.

\subsection{Action Run Frequency}
Action run frequency is computed for every action and subtask in a task. It records the number of runs in which a given action or subtask was used.

\subsection{Action Term Frequency}
Action term frequency is computed for every action and subtask in a task. It records the total number of times a given action or subtask was used in the task.

\subsection{Percentage Of Users In A Strategy}
The percentage of users in a strategy is computed for every BoT and echo strategy in a task. This records the percentage of total users that are grouped within the strategy. The sum of all BoT or echo strategy percentages would equal 100\%. The sum of all echo strategies in a BoT strategy would equal the percentage of users in the BoT strategy.

\subsection{Descriptions \& Side Effects}
Descriptions and side effects are recorded for every run and subtask in a task. A description is a manually defined description of what an action does, written as if it was preceeded by the phrase ``the user attempted to...''. For example, the description for ``hydra'' is ``use parallelized brute force password cracking on an online network protocol''. When computed for a run or subtask, all descriptions for each action present are gathered into a single string description that describes what the user did.

A side effect is a manually defined consequence of what an action does. For example, the side effect for ``hydra'' is ``access gain to online network protocol''. When computed for a run or subtask, all side effects for each action present are gathered into a string of side effects that lists the consequences of the run/subtask.

\subsection{Success Rate}
Success rate is computed for every BoT and echo strategy in a task. It records the number of runs in the strategy that lead to a successful completion of the task.

\section{Task Explorer GUI} 

The Task Explorer application is a GUI application built with Rust's immediate-mode GUI library egui. The Task Explorer application was compiled into a standalone Windows executable that can be used on any modern Windows operating system. The Task Explorer application allows users to easily visualize the results and statistics produced by the task explorer pipeline.

\subsection{Data Input/Selecting a New Directory}
\begin{figure}[h]
\begin{center}
\includegraphics[width=11cm]{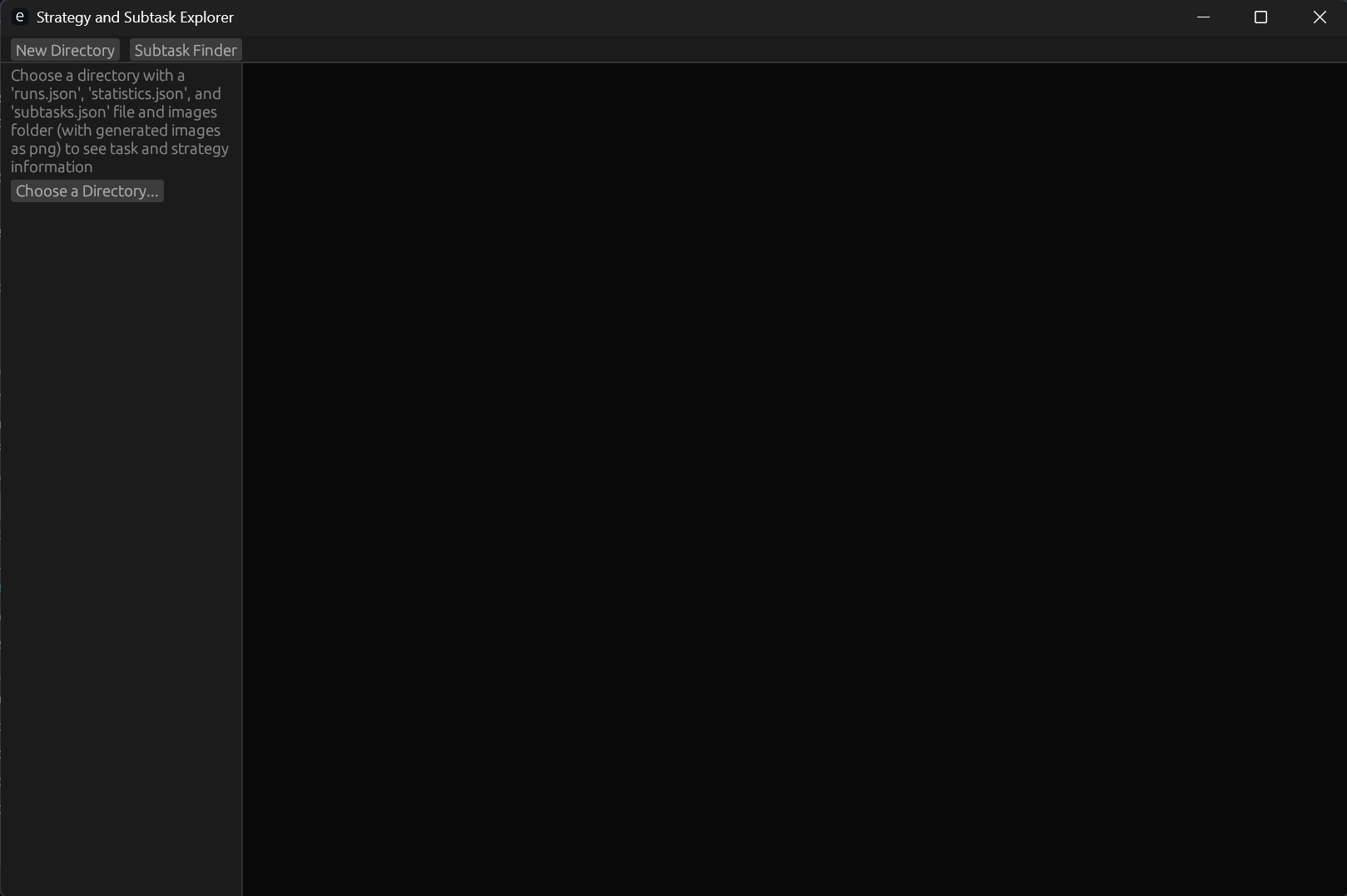}
\caption{Selecting A New Directory In The Task Explorer Application}
\label{newDirectory}
\end{center}
\end{figure}

Upon opening the Task Explorer application, the user will have to select a new directory by selecting the ``New Directory'' tool and then clicking ``Choose a Directory'' 
 (Figure \ref{newDirectory}). The user can then navigate to and select a folder that contains the \textit{statistics.json}, \textit{runs.json}, and \textit{subtask.json} artifact files, along with an images folder that contains all artifact images. These artifacts are generated from a Python script that deploys the task explorer pipeline on a set of tasks that are completed within an event. A new directory can be chosen at any time to load new results.

\subsection{The Four Quadrants}
\begin{figure}[h]
\begin{center}
\includegraphics[width=11cm]
{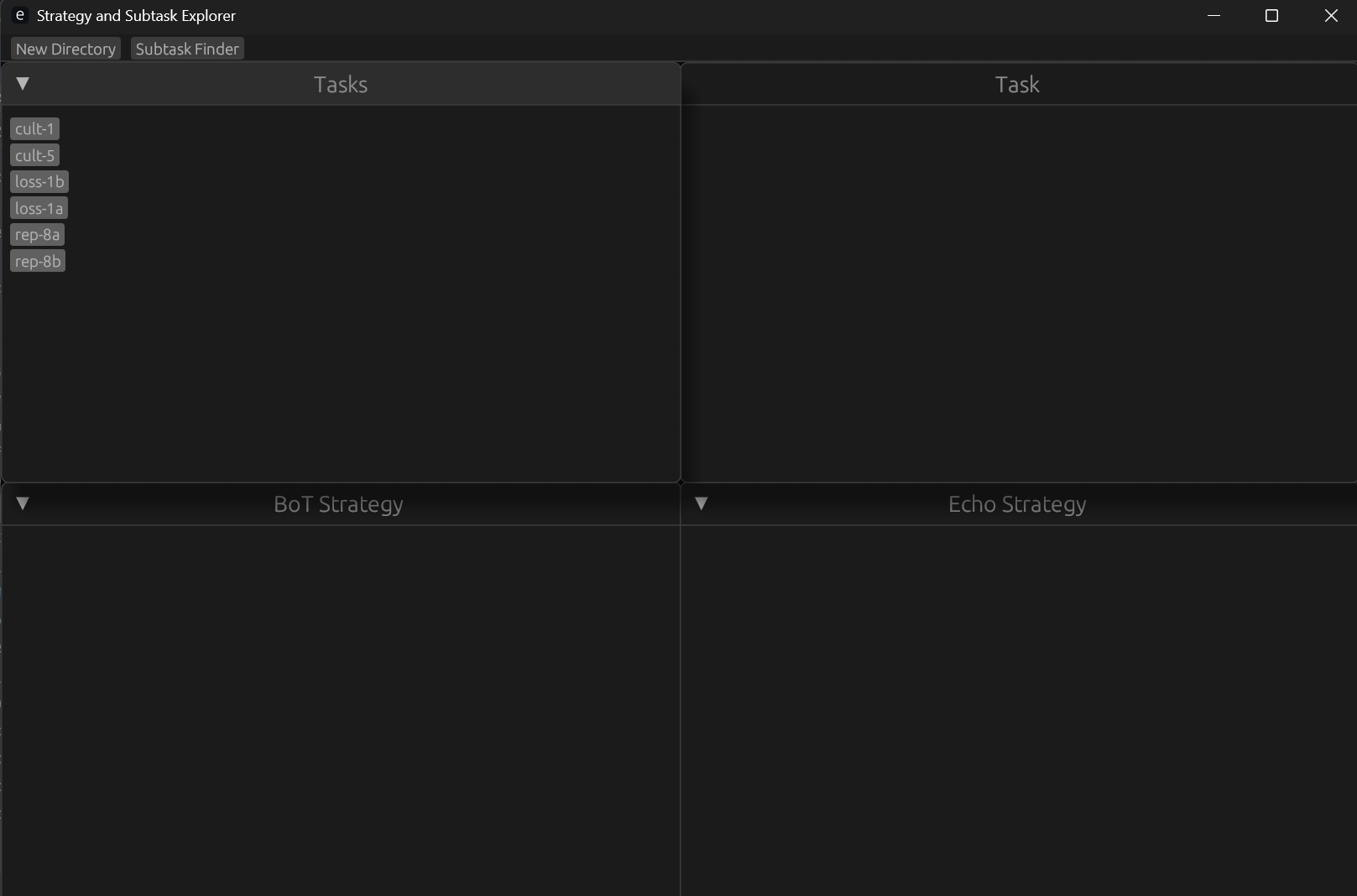}
\caption{The Four Quadrants And Tasks Panel}
\label{tasksPanel}
\end{center}
\end{figure}

After reading the artifact data, the Task Explorer presents four quadrants (Figure \ref{tasksPanel}): a Tasks panel, a Task panel, a BoT Strategy panel, and an Echo Strategy panel. These panels are meant to guide the user through the hierarchical nature of the task explorer pipeline's results. Users start in the Tasks panel, then move to the Task panel, then to BoT Strategy, and finally to Echo Strategy.

\subsection{Tasks Panel}
The Tasks panel (Figure \ref{tasksPanel}) displays all tasks that the task explorer pipeline ran on in a given event. Users can click on a task to select it, which updates the Task panel with the selected task's results.

\subsection{Task Panel}
\begin{figure}[h]
\begin{center}
\includegraphics[width=11cm]{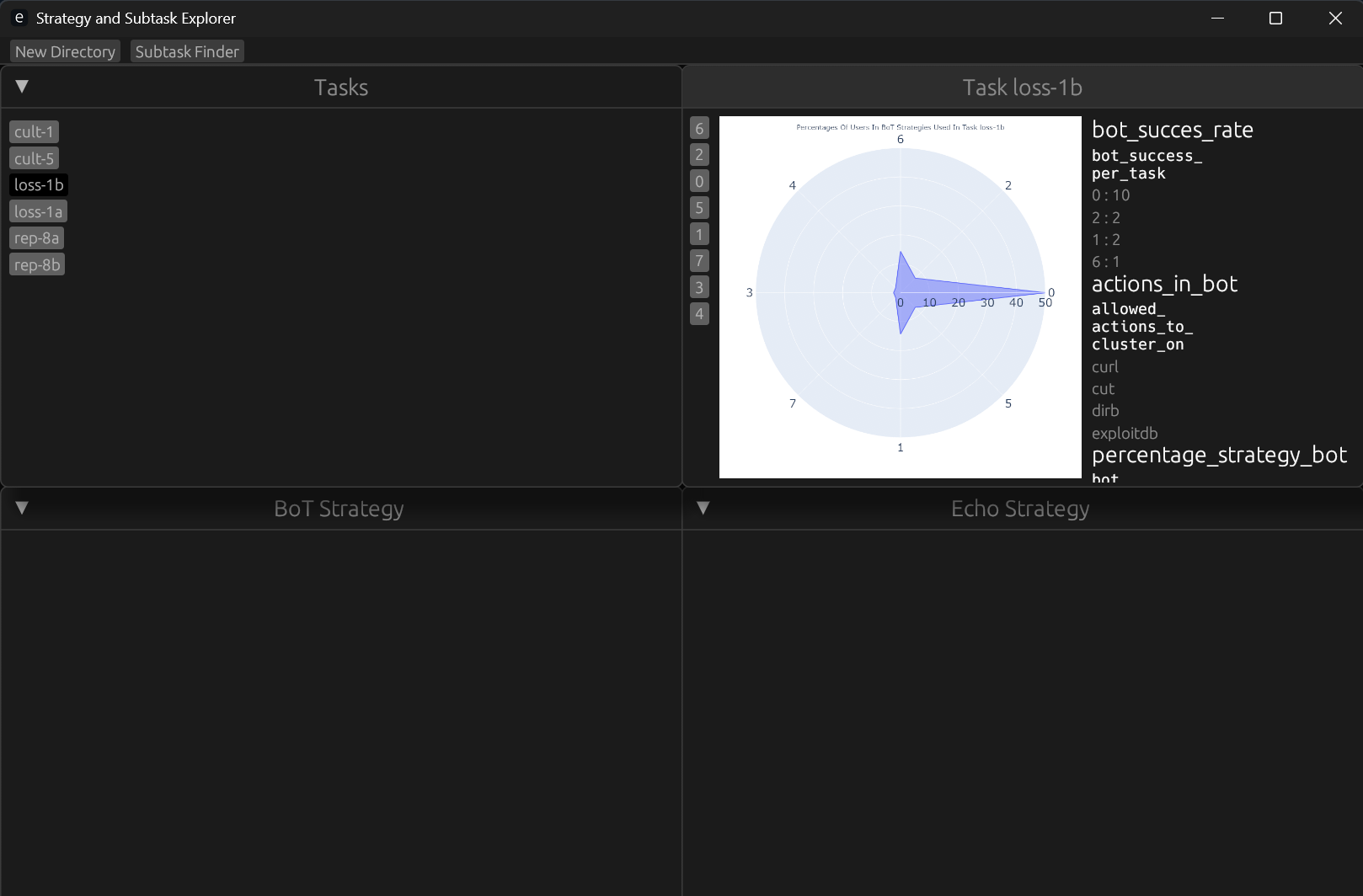}
\caption{The Task Panel}
\label{taskPanel}
\end{center}
\end{figure}

The Task panel (Figure \ref{taskPanel}) displays the BoT strategies that were identified within the selected task. Next to the BoT strategies, there is a spider graph to illustrate the percentage of users in each BoT strategy. If you scroll down on the image, the task's hierarchical clustering dendrogram will be just below the spider graph. To the right of these images is a scrollable area that presents any statistics that were computed for the selected task. Users can click on a BoT strategy to select it, which updates the BoT Strategy panel with the selected BoT Strategy's information. 

\subsection{BoT Strategy Panel}
\begin{figure}[h]
\begin{center}
\includegraphics[width=11cm]{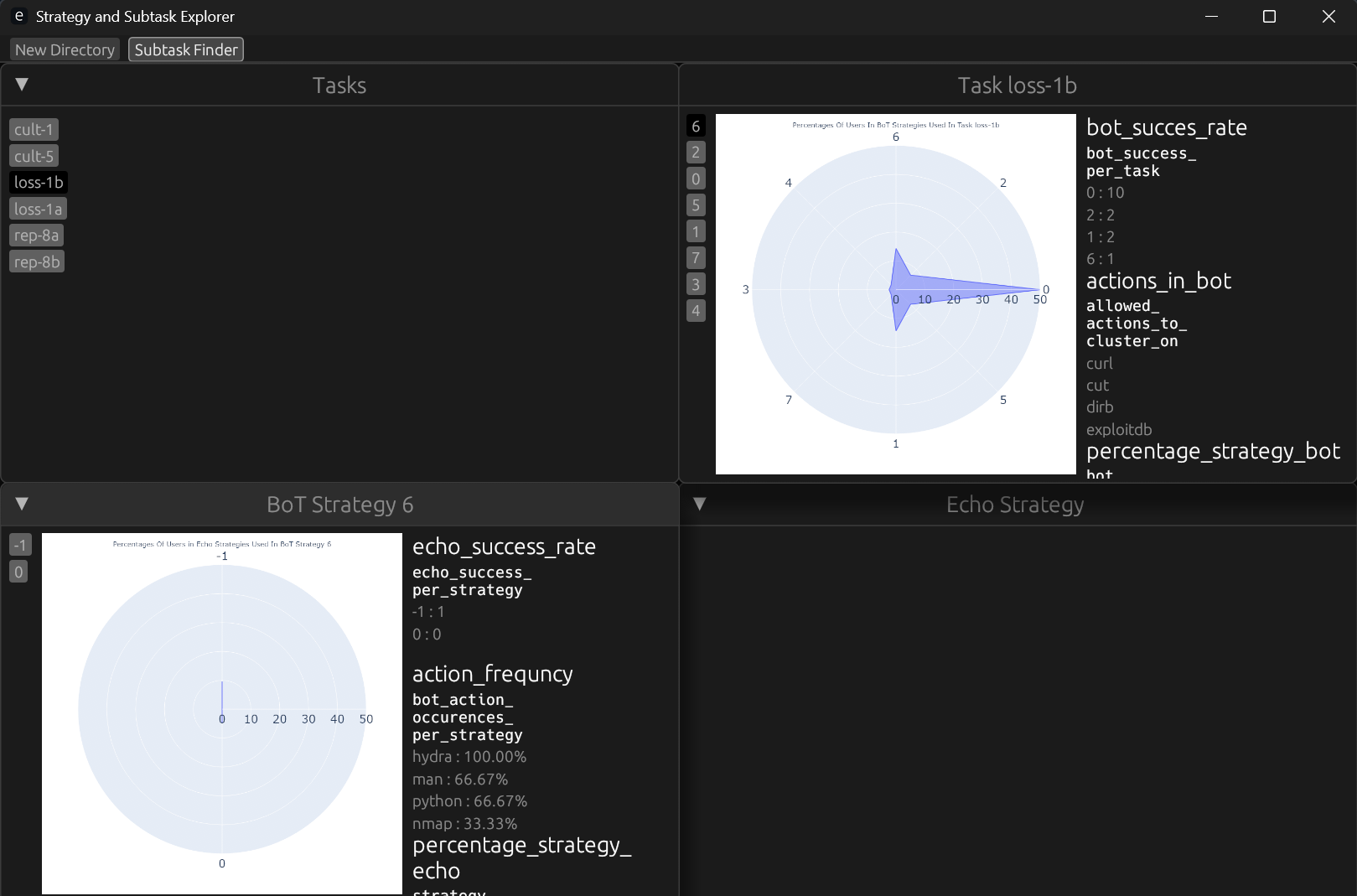}
\caption{The BoT Strategy Panel}
\label{botPanel}
\end{center}
\end{figure}

The BoT Strategy panel (Figure \ref{botPanel}) displays the Echo strategies that were identified within the selected BoT strategy. Next to the Echo strategies, there is a spider graph to illustrate the percentage of users in each Echo strategy. To the right of the graph is a scrollable area that presents any statistics that were computed for the selected BoT strategy. Users can click on an echo strategy to select it, which updates the Echo Strategy panel with the selected echo strategy's information. 

\subsection{Echo Strategy Panel}
\begin{figure}[h]
\begin{center}
\includegraphics[width=11cm]{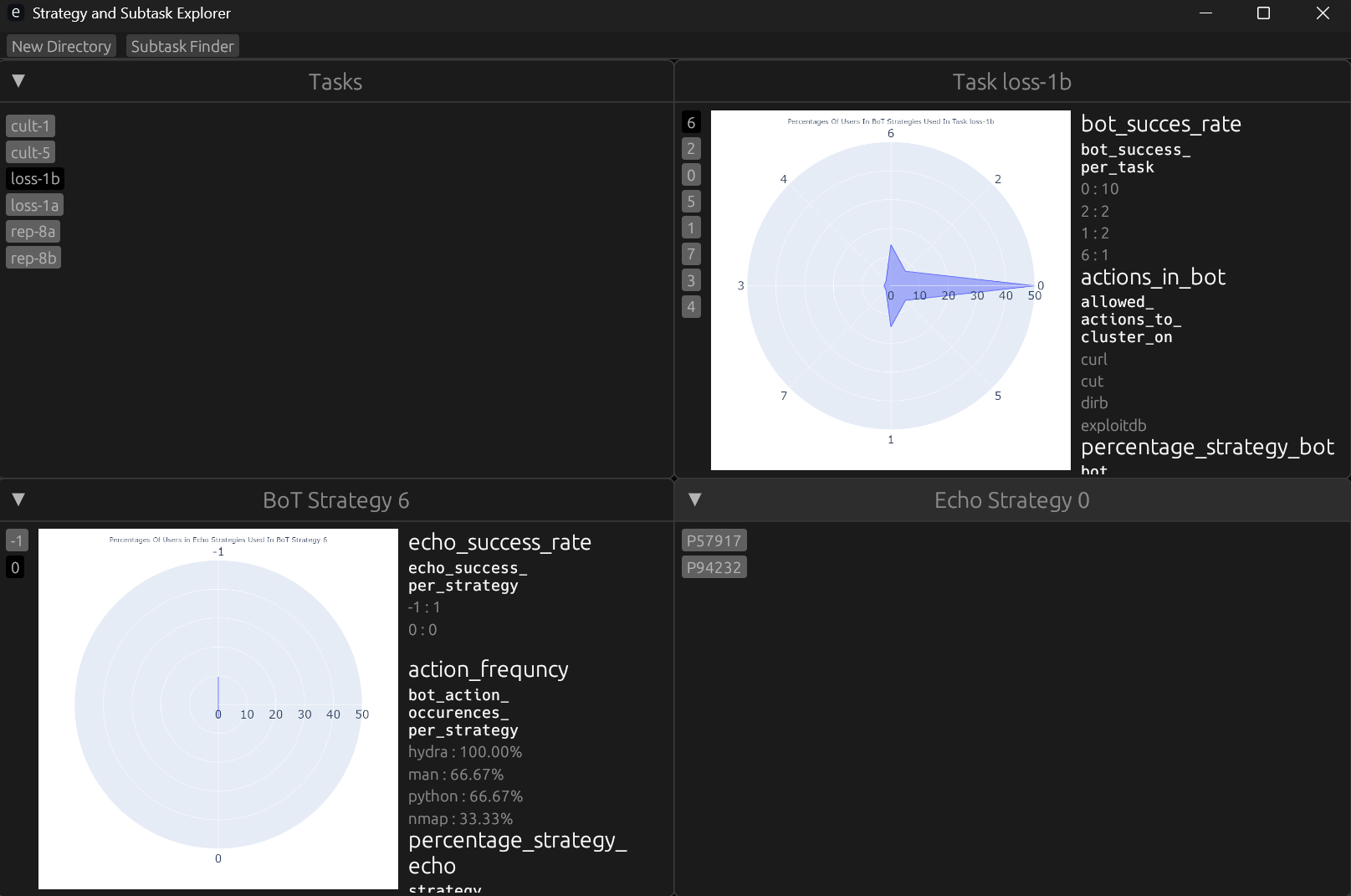}
\caption{The Echo Strategy Panel}
\label{echoPanel}
\end{center}
\end{figure}

The Echo Strategy panel (Figure \ref{echoPanel}) displays the user runs that were grouped within the selected echo strategy. Next to the user runs, any statistics that were computed for the selected echo strategy would show, but currently there are none. Users can click on a run to open the User pop-up for the selected user. 

\subsection{User Pop-Up}
\begin{figure}[h]
\begin{center}
\includegraphics[width=11cm]{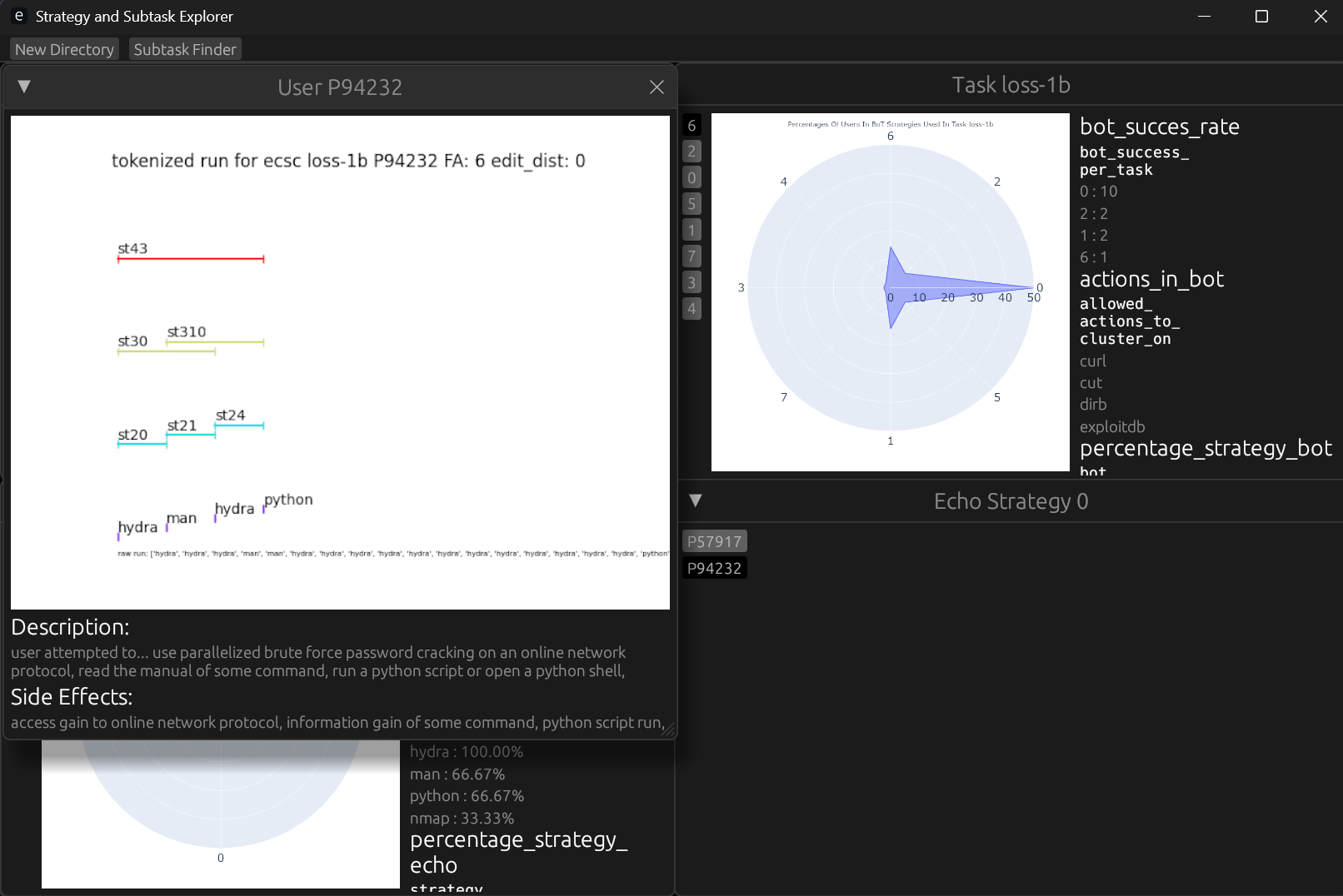}
\caption{The User Pop-Up}
\label{userPopUp}
\end{center}
\end{figure}

The User pop-up (Figure \ref{userPopUp}) displays the selected user's hierarchical encoding diagram. Under the hierarchical encoding diagram, the run's description and side effects are listed. 

\subsection{Subtask Finder Pop-Up}
\begin{figure}[h]
\begin{center}
\includegraphics[width=11cm]{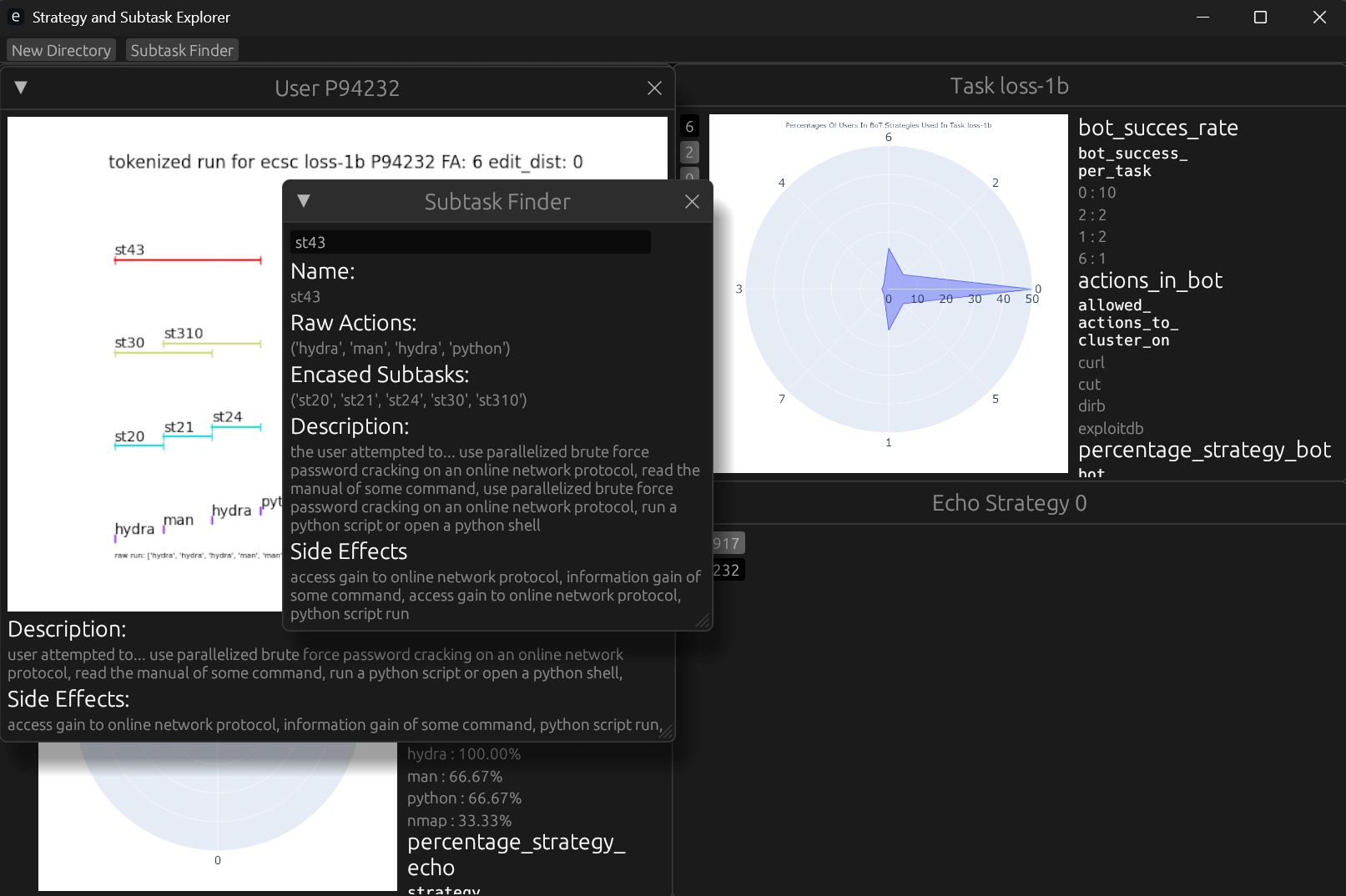}
\caption{The Subtask Finder Pop-Up}
\label{subtaskFinderPopUp}
\end{center}
\end{figure}

The Subtask Finder pop-up (Figure \ref{subtaskFinderPopUp}) can be summoned by clicking on the ``Subtask Finder'' tool at the top of the application. It allows the user to enter a search for any real subtask and get that subtask's name, raw actions, encased subtasks, description, and side effects.

\subsection{Other Features}

\begin{figure}[h]
\begin{center}
\includegraphics[width=11cm]{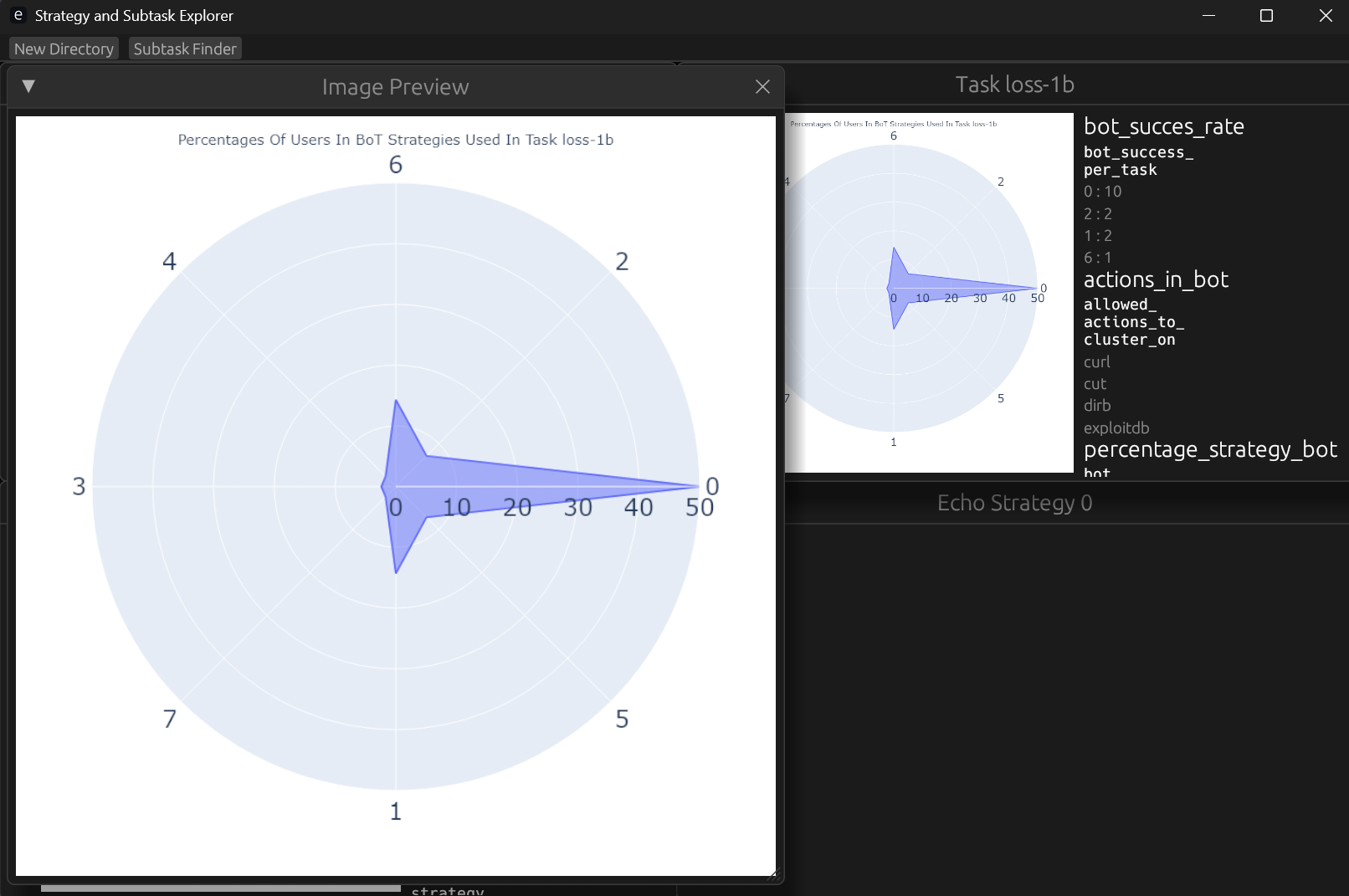}
\caption{The Subtask Finder Pop-Up}
\label{imagePreviewPopUp}
\end{center}
\end{figure}

If the user wants to focus on any displayed image, they can simply click on the image to summon the Image Preview pop-up (Figure \ref{imagePreviewPopUp}). This displays the selected image in a pop-up that can be dragged and resized to view the image at different sizes. The pop-up also retains the image when selecting a different portion of the application, letting the user compare the image in the pop-up to other images that were updated by new selections.

\begin{figure}[h]
\begin{center}
\includegraphics[width=11cm]{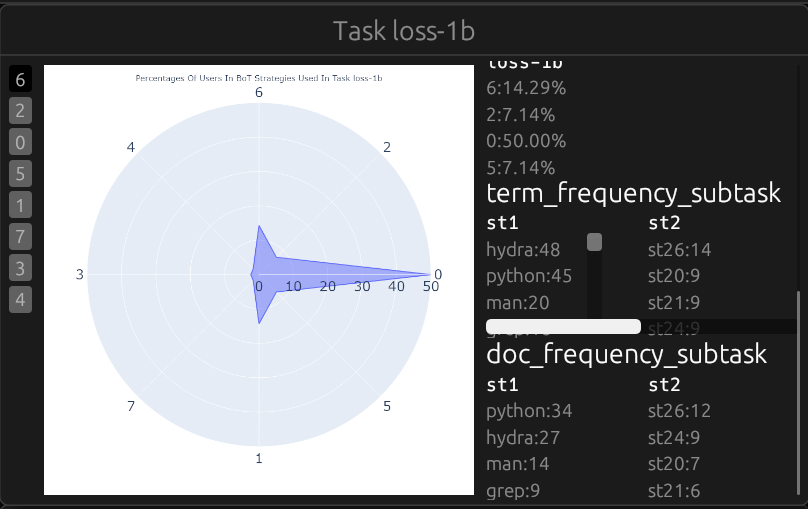}
\caption{An Example Of The Horizontal and Vertical Scrolling Functionality In A Statistic Instance}
\label{statisticsExample}
\end{center}
\end{figure}

The statistics section in every panel is built to be infinitely scalable, allowing an infinite amount of statistics to be displayed. This was done to allow new statistics to be easily added at any time, without needing to update the application. Every statistic has a statistic type and a sub-statistic type. Only one instance of a statistic type can be displayed per panel, but infinite amounts of vertically scrolling sub-statistics can be displayed in a horizontal scrolling section within the statistic instance. Figure \ref{statisticsExample} shows an example of this scalable design, where ``term\_frequency\_subtask'' is the statistic instance of the ``Action Term Frequency'' statistic defined earlier. Inside of the statistic instance, there is a horizontal scrolling area for each sub-statistic type. The sub-statistics for term\_frequency\_subtask are the n-gram sizes for each subtask size. For each sub-statistic, there is a vertically scrolling instance listing each action/subtask and their term frequency.

To add more statistics, one must simply add the statistics for each statistic instance to the \textit{statistics.json} artifact file using the task explorer pipeline python script. An example statistic can be seen in Figure \ref{statisticJsonExample}. From top to bottom, ``task'' tells the application panel to display the statistic in the Task panel, ``term\_frequency\_subtask'' is the unique statistic name, ``st2'' is the sub-statistic name, ``loss-1b'' is the task to display the statistic for, ``st214:14|st20:9|st21:9 ...'' is the statistics separated by the bar character, and ``Subtask Term Frequency'' is the user-friendly name of the statistic.

\begin{figure}
\begin{center}
\begin{tabular}{ |l|}
    \hline
    \{\\
        ``hierarchyLevel'': ``task'',\\
        ``statType'': ``term\_frequency\_subtask'',\\
        ``statSubtype'': ``st2'',\\
        ``identifier'': ``loss-1b'',\\
        ``statsList'': ``st214:14|st20:9|st21:9 ...'',\\
        ``header'': ``Subtask Term Frequency''\\
    \}\\
    \hline
\end{tabular}
\caption{An Example Of A Single Statistic Instance In The \textit{statistics.json} Artifact File}
\label{statisticJsonExample}
\end{center}
\end{figure}

\chapter{Conclusions}

Creating machine systems that can anticipate user needs and behaviors is becoming increasingly relevant in our information-centric society. Scholars indicate that aHMI is a new field that needs to be further explored to create anticipatory machines, especially with respect to implicit human-machine coordination. This research created an automated task exploring pipeline that identifies key strategies at two levels: A global set of strategies that gathers distinct sets of actions used to complete a task and a local intra-cluster set of strategies where user runs use those actions in a similar way compositionally. Additionally, the task explorer pipeline defines meaningful subtasks that are hierarchically combined and linked together to identify common micro user traces. Strategies and subtasks help inform both humans and machines about a user's knowledge, skills, and behaviors.

The primary limitation to this study is the length of most user run time-series data in the CTF dataset. Most user runs were rather short, whereas many other real-world examples require larger lengths of actions and may not be as easily clustered. The other limitation to this research is the maximum length of subtasks. Subtasks were restricted to having two to four actions. Although this works well for the CTF dataset, other real-world examples would likely find larger subtasks as the length of actions needed to complete a task increases.

This research is a first step in advancing human-machine coordination, by identifying common strategies and meaningful subtasks used to complete a task. Future research would include parameter generalization and action linking. The current implementation looks at only fundamental actions without parameters, but including generalized parameters can give additional context to each action and can lead to even more meaningful connections between actions. There are likely general links where an action being executed with one set of parameters leads to a proceeding action call, while the same action with a different set of parameters leads to a separate action call. Additionally, future work would incorporate strategy and task prediction based on the subsequence of a user run.

While our task explorer pipeline was deployed on CTF commands, the pipeline can be easily modified to fit any dataset that is comprised of user run time-series data. For any given task, the pipeline will help inform humans and machines about the typical knowledge, skills, and behaviors required to complete the task. This information is valuable because once the machine identifies what the user's knowledge of a task is and how the user typically completes similar tasks, the machine can start to anticipate the user's future behaviors with implicit coordination.
%
% ***** bibliography/references in file bib.tex
%
%\input bib.tex
% Do not use bibtex if you plan to use references for the URI-derived thesis template.
% Instead, you should reference the file "uribibtex.bat" instead of the standard bibtex
\bibliographystyle{ieeetr}
\reffile{references} % To build the references, double click the file "genbib.bat" in the "build" subdirectory

%
% ***** appendices in appA.tex, appB.tex, ...
%

\include{appA}

\end{document}